\pdfoutput=1

\documentclass[11pt]{article}

\usepackage[final]{acl}
\usepackage{lipsum}
\usepackage{times}
\usepackage{latexsym}
\usepackage{amsmath}

\usepackage{booktabs}
\usepackage{float}
\usepackage{caption}
\usepackage{footnote}
\makesavenoteenv{figure}

\usepackage[T1]{fontenc}

\usepackage[utf8]{inputenc}
 \usepackage{booktabs}

\usepackage{microtype} 

\usepackage{inconsolata}

\usepackage{graphicx}
\usepackage{xcolor}
\usepackage{soul}

\definecolor{lightred}{rgb}{1.0, 0.8, 0.8}

\title{ShadowLLM: Predictor-based Contextual Sparsity\\for Large Language Models}

\author{
\setcounter{footnote}{1}
    Yash Akhauri\textsuperscript{\rm{1},\rm{2}}, 
    Ahmed F. AbouElhamayed\textsuperscript{\rm{1}}, 
    Jordan Dotzel\textsuperscript{\rm{1},\rm{2}}, 
    Zhiru Zhang\textsuperscript{\rm{1}}, \\
    \textbf{Alexander M. Rush}\textsuperscript{\rm{1}}, 
    \textbf{Safeen Huda}\textsuperscript{\rm{2}}, 
    and \textbf{Mohamed S. Abdelfattah}\textsuperscript{\rm{1}} \\
    \textsuperscript{1}Cornell University \textsuperscript{2}Google \\
    \texttt{\{ya255, afa55, jad443\}@cornell.edu}\\
    \texttt{\{zhiruz, arush, mohamed\}@cornell.edu},\;\;\texttt{safeen@google.com} \\
}

\usepackage{tikz}
\usepackage{xcolor}

\begin{document}

\maketitle
\begin{abstract}
The high power consumption and latency-sensitive deployments of large language models (LLMs) have motivated efficiency techniques like quantization and sparsity. 
\textit{Contextual sparsity}, where the sparsity pattern is input-dependent, is crucial in LLMs because the permanent removal of attention heads or neurons from LLMs can significantly degrade accuracy. 
Prior work has attempted to model contextual sparsity using neural networks trained to predict activation magnitudes, which can be used to dynamically prune structures with low predicted activation magnitude.
In this paper, we look beyond magnitude-based pruning criteria to assess attention head and neuron importance in LLMs.
We develop a novel predictor called ShadowLLM, which can \textit{shadow} the LLM behavior and enforce better sparsity patterns, resulting in over 15\% improvement in end-to-end accuracy compared to prior methods. 
In addition, ShadowLLM achieves up to a 20\% speed-up  over the state-of-the-art DejaVu framework. These enhancements are validated on Llama-2 and OPT models with up to 30 billion parameters. Our code is available at \href{https://github.com/abdelfattah-lab/shadow_llm/}{ShadowLLM}.
\end{abstract}


\begin{figure}
    \centering
    \includegraphics[width=\columnwidth]{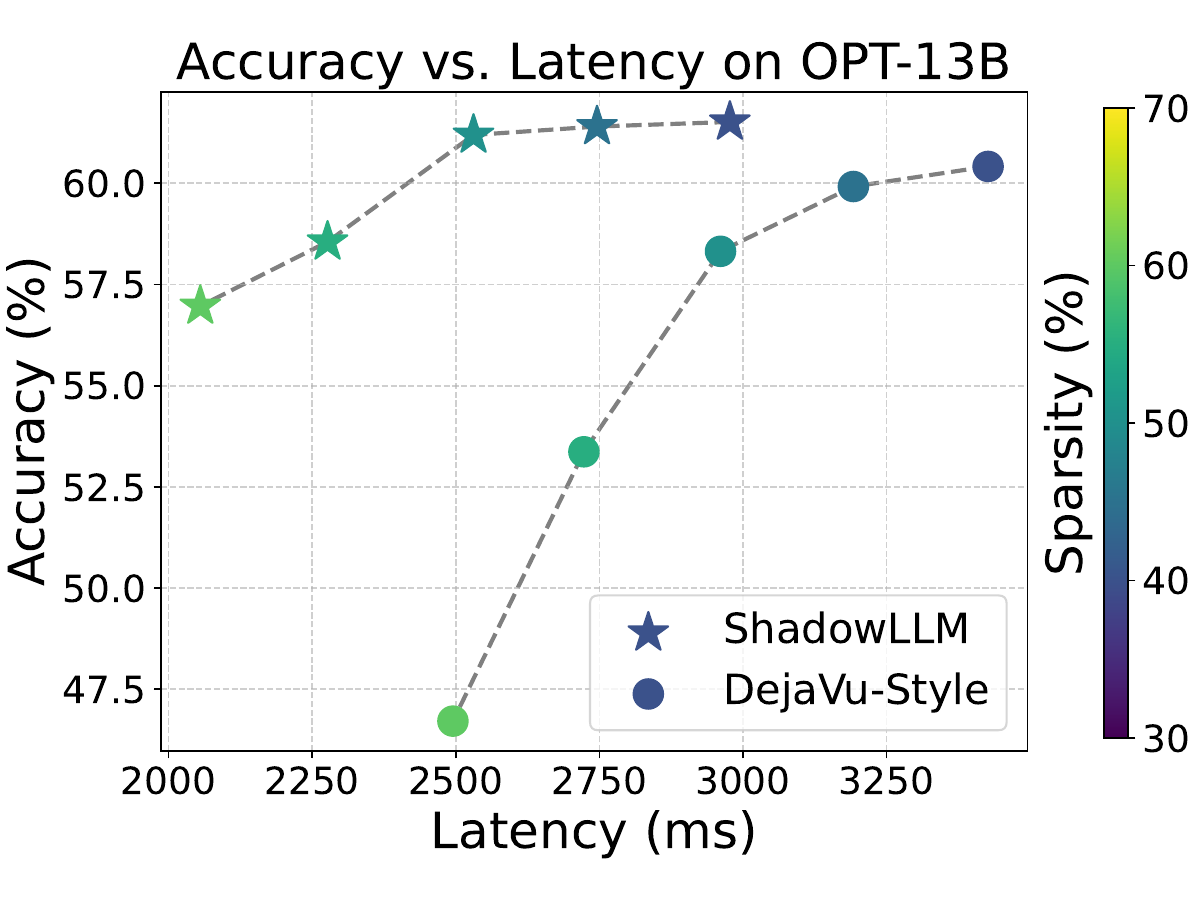}
    \caption{ShadowLLM uses more accurate pruning criteria and a simpler sparsity predictor compared to DejaVu. Its pruning criteria results in a stronger accuracy-sparsity trade-off (geomean) across seven downstream evaluation tasks, and its unified predictor improves the execution latency compared to the layerwise predictor of DejaVu.}
    \label{fig:teaser_fig}
\end{figure}
\section{Introduction}

\begin{figure*}[ht]
    \centering
    \begin{minipage}{0.49\textwidth}
        \centering
        \includegraphics[width=\textwidth]{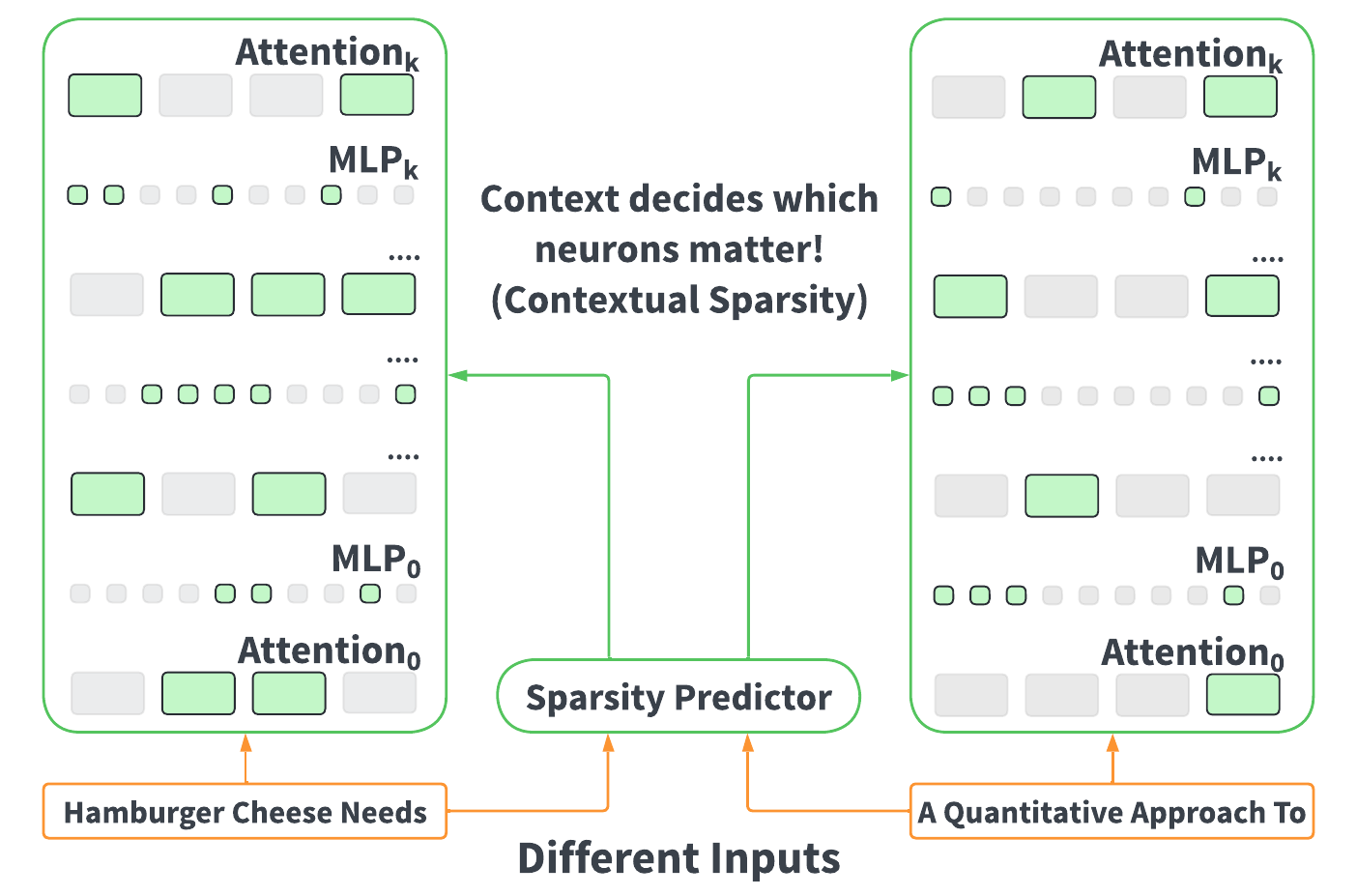}
        \caption{Contextual sparsity prunes neurons and attention heads based on the context (input) itself. Training a predictor to dynamically predict the sparsity pattern dependent on the input tokens can improve model quality.}
        
        \label{fig:predictor_concept}
    \end{minipage}
    \hfill
    \begin{minipage}{0.49\textwidth}
    \centering
    \includegraphics[width=\columnwidth]{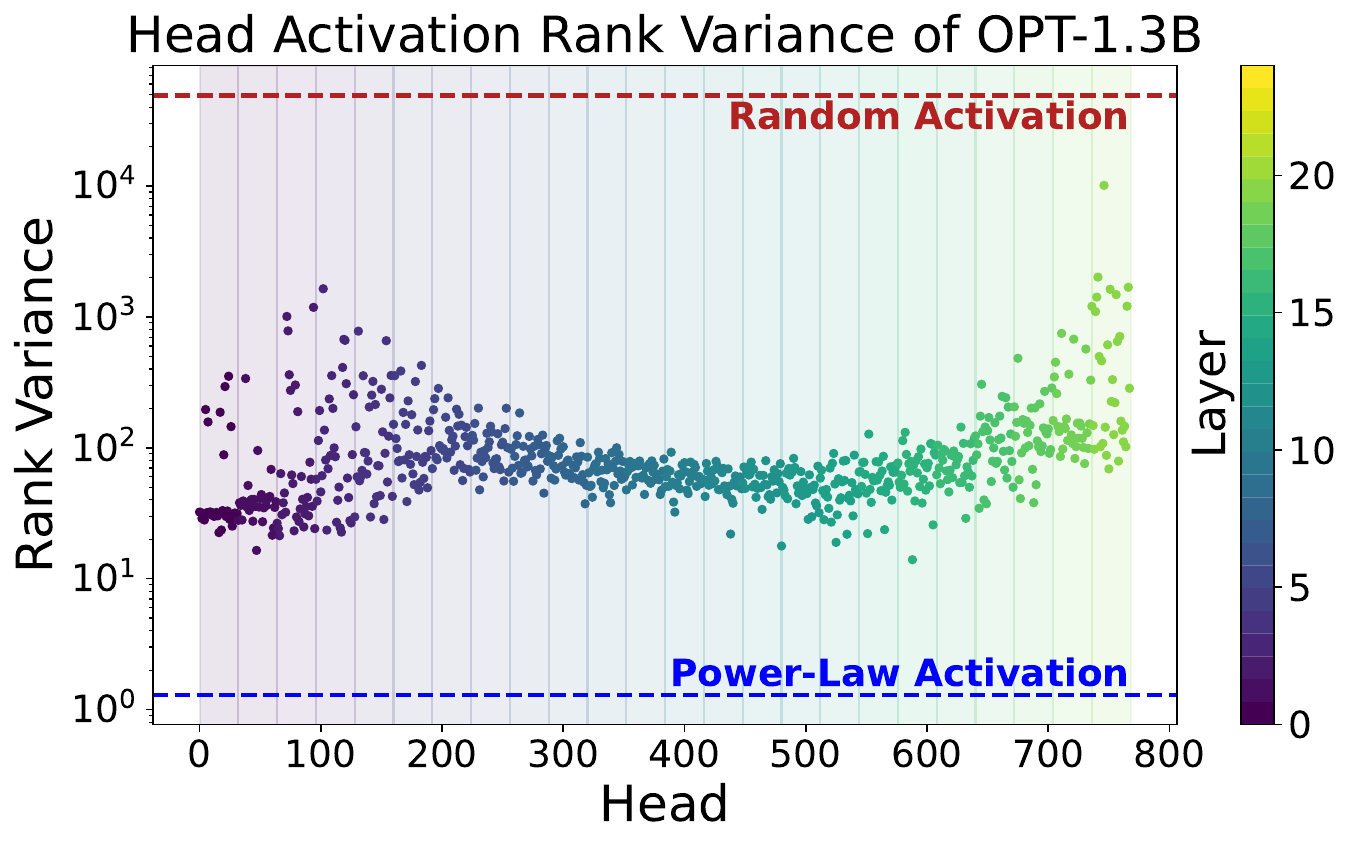}
    \caption{Heads with higher rank variance, calculated using \texttt{GradNorm}, indicate greater context dependence. This context dependence, or contextual sparsity, is most noticeable in the early and later layers of the OPT-1.3B model. We measured the variance in rank for each head across 5000 inputs in seven five-shot evaluation tasks.}
    \label{fig:powerlaw_ablation}
    \end{minipage}
\end{figure*}

Large language models (LLMs) are emerging as a core component of many computing applications. 
Their ability to perform in-context learning, i.e., to perform a task by conditioning on examples without any gradient updates~\cite{brown2020language, liang2022holistic, min2022rethinking}, make them broadly applicable to numerous applications. 
Yet, their large size combined with the latency-sensitivity of LLM-based applications make them expensive to deploy~\cite{hoffmann2022training}. 

A key optimization in LLM deployment is sparsification, where weights or activations are pruned to reduce the computation and memory requirements at run time. 
Sparsification can either be \textit{static}, which permanently removes an attention head or neuron, or \textit{contextual}, which prunes based on the current input and context.
While some works investigate task-specific static pruning methods~\cite{Bansal2022RethinkingTR, michel2019sixteen}, they typically have a large impact on in-context learning, reducing downstream task accuracy compared to contextual sparsity.

Contextual sparsity can be leveraged at run time to dynamically prune LLMs, yet it requires making fast and accurate predictions based on predetermined \textit{pruning criteria}. 
These criteria can have large effects on the overall accuracy and performance of the model, as shown in Figure~\ref{fig:teaser_fig}.
Our method ShadowLLM uses more accurate pruning criteria and a unified predictor at the beginning of the model, which leads to a stronger accuracy-performance tradeoff compared to the recent work DejaVu~\cite{liu2023deja}. 

Both of these methods dynamically vary their sparsity patterns given different inputs using sparsity predictors, as shown in Figure~\ref{fig:predictor_concept}.
The inputs are passed into a sparsity predictor, which then outputs the corresponding per-layer masks on the attention and MLP layers.
For DejaVu, the sparsity pattern is generated with neural-network predictors at each layer.
This gives access to more local information, but layerwise predictors come with an expensive run-time cost.

On the model quality side, contextual sparsity exists if there is a significant variance on head and neuron importance as the input changes. 
Figure \ref{fig:powerlaw_ablation} quantifies this variance on the importance (ranks) of attention heads on OPT-1.3B across different inputs.
It demonstrates the relative importance changes significantly, especially in the earlier and the later layers. Naturally, this variance across inputs necessitates a dynamic pruning strategy to ensure an appropriate quality--latency trade-off. 

In this work, we explore the effects of different pruning criteria and predictor design on LLM accuracy and latency. Our contributions are summarized below:
\begin{enumerate}
    \item \textbf{Pruning Criteria:} We evaluate approaches from prior pruning research to find head and neuron pruning criteria that can improve downstream zero-shot accuracy by 15\% without affecting performance.
    \item \textbf{Early Prediction:} We use a single predictor at the first layer of the LLM to model the entire LLM sparsity pattern, improving performance by 20.6\% without affecting accuracy.
\end{enumerate}

\section{Related Work}

\subsection{Pruning Criteria}
Research in the area of designing criteria for pruning neurons has focused on using the activations, weights, and gradients of neural networks to assess the relative importance of neurons. Several pruning criteria have been designed to utilize light-weight computations, such as a single forward-backward pass through the network. For instance, some works use parameter magnitudes as a proxy for parameter saliency~\cite{frankle2018lottery, han2015learning}, whereas others use gradient-based information~\cite{NIPS1989_6c9882bb, hassibi1992second, molchanov2016pruning, Bansal2022RethinkingTR}. Further, research in Neural Architecture Search (NAS) adapts these pruning criteria to assess and compare entire architectures. Such initialization-based measures like NASWOT~\cite{mellor2021neural} aim to study other properties of the architecture, and can be used to study neuron importance as well.

In this work, we adapt several neuron importance criteria from research in pruning and NAS~\cite{abdelfattah2021zerocost, Lopes_2021, mellor2021neural, turner2019blockswap} to evaluate which methods work well for dynamic pruning of large language models at run time.

\subsection{LLM Inference Optimization}
Given the recent exponential increase in model size, significant research has been dedicated to optimizing NN inference to decrease compute, power, and latency. 
Quantization reduces the precision of model parameters, embeddings, and key-value caches~\cite{zhang2023binarized, dotzel2024students, zhao2024atom}.
Orthogonal to quantization, there has been research on accelerating sparse language models, which either statically or dynamically trim portions of compute throughout the model~\cite{hua2019channel, schuster2022confident, elbayad2020depthadaptive}. 

Within these works, DejaVu~\cite{liu2023deja} leverages dynamic sparsity by building predictors to estimate sparsity patterns. In this paper, we investigate how the predictor can be improved, both in terms of performance and model quality.


\begin{figure}[t!]
    \centering
    \includegraphics[width=\columnwidth]{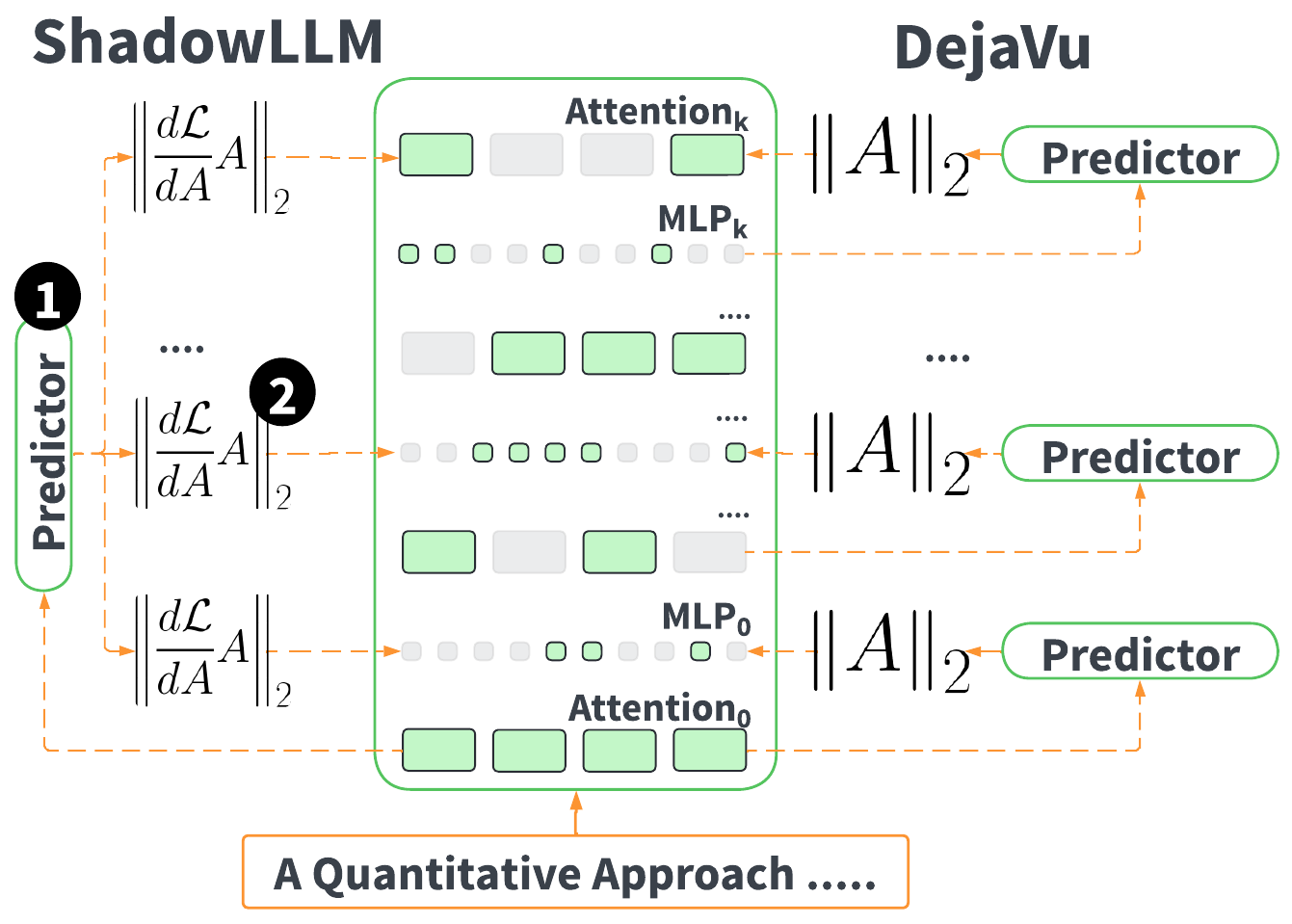}
    \caption{\textbf{(1)} A single predictor to model the entire LLM improves model performance, while \textbf{(2)} utilizing gradient based information when evaluating pruning criteria for neurons improves model quality.}
    \label{fig:methodology_components}
\end{figure}

\section{Pruning Criteria}
\label{sec:pruning_criterion}

Contextual sparsity requires dynamically understanding which neurons to prune (i.e. assessing the neurons importance relative to an input) and ranking the neurons relative to each other.
Figure \ref{fig:methodology_components} depicts how we can use information about the activations and gradients to prune a LLM for this contextual sparsity.

Consider a model $\mathcal{M}$ and dataset $\mathcal{D}$, containing prompts (inputs) along with the target output sequence. We then wish to define performance on the dataset as $\mathcal{P}_{\mathcal{M}}(\mathcal{D})$. Now suppose a subset of the model $\mathcal{C} \subset \mathcal{M}$ is pruned out. Ideally, we would like to be able to estimate $\mathcal{P}_{\mathcal{M}}(\mathcal{D}) - \mathcal{P}_{\mathcal{M} \backslash \mathcal{C}}(\mathcal{D})$ ~\cite{Bansal2022RethinkingTR}.

The optimal pruning strategy is found in Equation \ref{eq:optpruned}. If we look at aggressive attention head pruning of even small transformers (prune 56 out of 64 heads in each layer), exhaustive search in a single layer would require $^{64}C_{8}$ evaluations, and this would have to be repeated for every layer, making the problem intractable. 

\begin{equation}
\arg \min_{\mathcal{C} \subset \mathcal{M}} \;\; \mathcal{P}_{\mathcal{M}}(\mathcal{D}) - \mathcal{P}_{\mathcal{M} \backslash \mathcal{C}}(\mathcal{D})
\label{eq:optpruned}
\end{equation}


We can feed a subset of the dataset $d \in \mathcal{D}$ to the model $\mathcal{M}$, and calculate the loss $\mathcal{L}$. Further, we can also get access to the activations ($A$), as well as the parameters of the up-projection FFN of transformer at layer $l$ as $\theta_{l}$. The activation at layer $l$ for the $k^{\text{th}}$ head or neuron is denoted as $A_{l,k}$. The gradients for these activations are denoted as $\frac{\partial \mathcal{L}}{\partial A_{l,k}}$. The gradient for the weight parameters of the $k^{\text{th}}$ neuron in the up-projection FFN at layer $l$ is given as $\frac{\partial \mathcal{L}}{\partial \theta_{l,k}}$.

\begin{figure*}[ht]
    \centering
    \begin{minipage}{0.49\textwidth}
        \centering
        \includegraphics[width=\columnwidth]{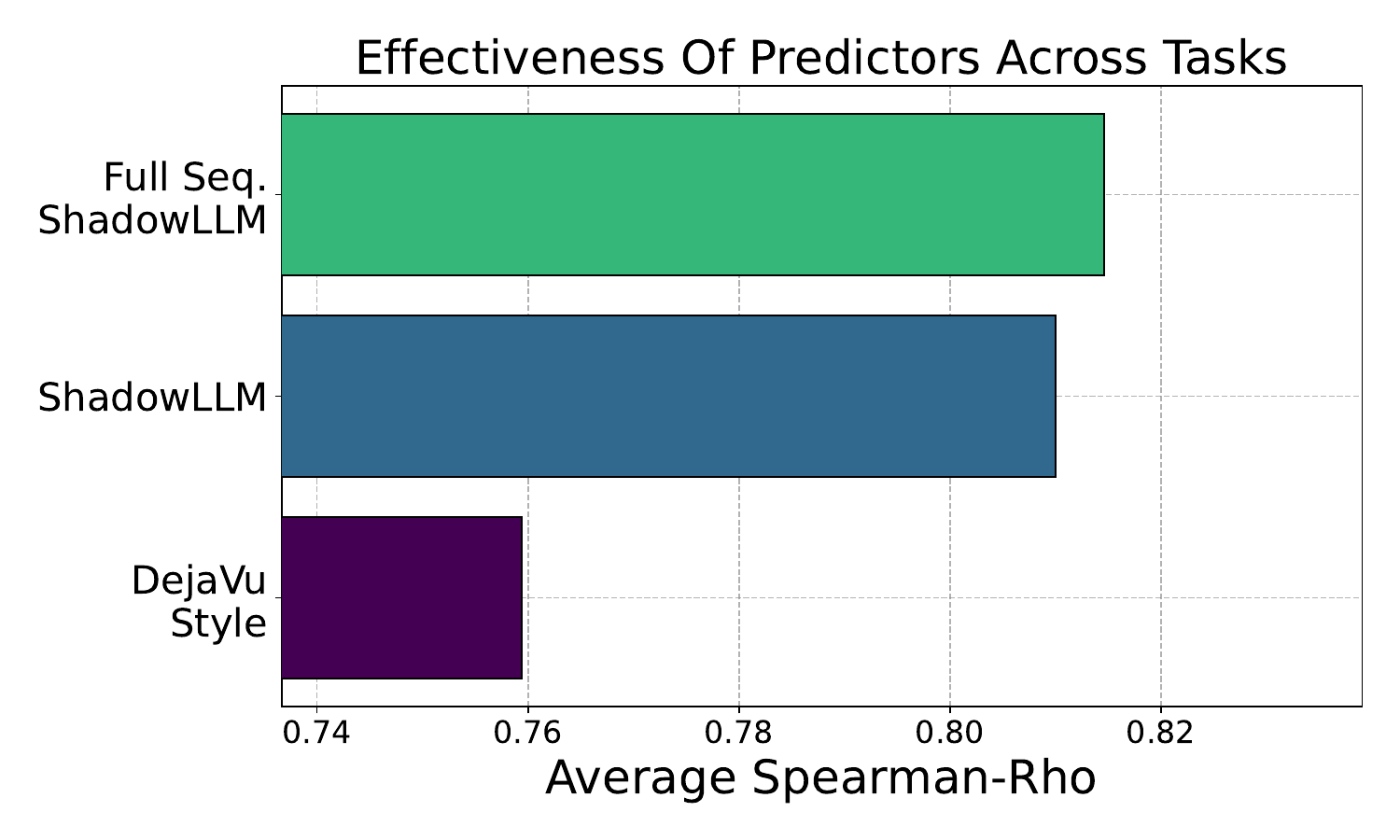}
        \caption{Head importance ranking ability of different sparsity predictors on 500 queries across 7 downstream tasks. A single predictor at the start of the transformer can accurately model the global relative head and neuron importance. }
        \label{fig:predictor_design_test}
    \end{minipage}
    \hfill
    \begin{minipage}{0.49\textwidth}
        \centering
        \includegraphics[width=\columnwidth]{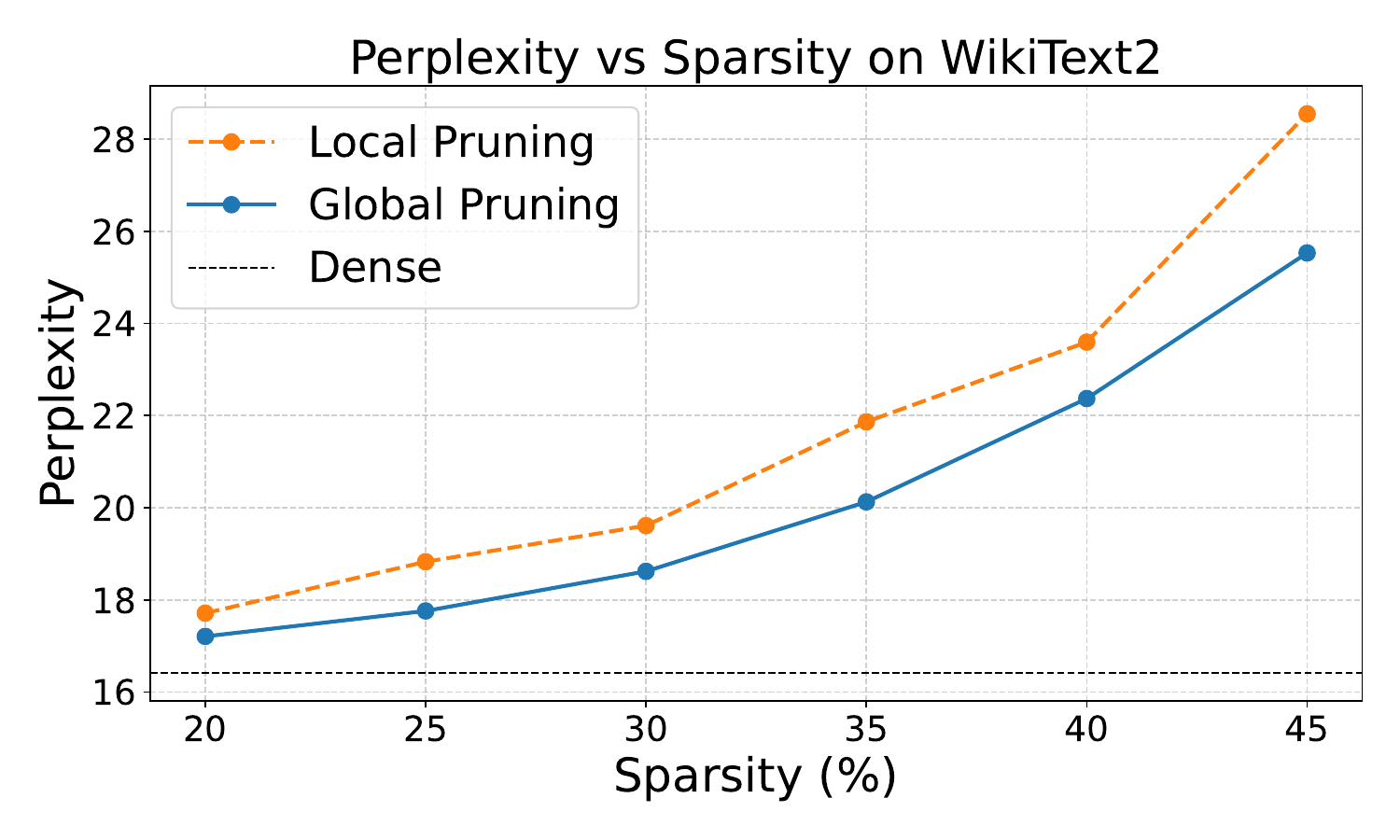}
        \caption{Global pruning outperforms local (per-layer) pruning strategies using ShadowLLM trained on the \texttt{plainact} criteria (OPT-1.3B). Global pruning accommodates the varying importance of different layers, allowing for unbalanced pruning across layers.}
        \label{fig:loc_glob_1.3b}
    \end{minipage}
\end{figure*}

Current predictor-based sparsity research investigates the impact of magnitude-based criteria, such as the \texttt{L2Norm} of the head and neuron activation on a subset of data $d$. The intuition is that the heads that are more \textit{activated} should be more important. There is significant research on other criteria for pruning weights and activations~\cite{molchanov2016pruning}. Beyond activation magnitude being a criterion for importance, the process of pruning can also be framed as an optimization problem, with the goal of approximating the change in loss from removing parameters. Methods such as optimal brain damage (OBD)~\cite{NIPS1989_6c9882bb} rely on the gradient of the loss with respect to the feature maps. While OBD evaluates the second-order term (Hessian), works such as ~\cite{figurnov2016perforatedcnns, molchanov2016pruning} come up with similar metrics based on the Taylor expansion of the change in loss.

In this paper, we evaluate pruning criteria of varying complexity, that use \textbf{(1)} Activation Methods, \textbf{(2)} First-Order Gradient (Jacobian) Methods, \textbf{(3)} Activation + Jacobian Methods, \textbf{(4)} OBD-style Hessian Methods and \textbf{(5)} Sensitivity-Based Methods for pruning LLMs. 

Among these methods, we find that a gradient-based sensitivity method we call \texttt{plainact} out-performs activation-based magnitude methods adapted in prior dynamic pruning research~\cite{liu2023deja}. The \texttt{L2Norm} activation-magnitude based criterion assesses the importance of neurons by simply taking the L2 Norm of the head and neuron activation as $|| A_{l,k} ||_2$. The \texttt{plainact} criterion measures the expected sensitivity of the model on the loss if a head or neuron is removed. For the head and neuron, this can be described simply as $||A_{l,k} \cdot \frac{\partial \mathcal{L}}{\partial A_{l,k}}||_{1}$ and $||\theta_{l,k} \cdot \frac{\partial \mathcal{L}}{\partial \theta_{l,k}}||_{1}$ respectively. We perform in-depth ablations across several pruning criteria in Section \ref{sec:analysis}, and find that \texttt{plainact} empirically performs well as a pruning criteria.

\section{Predictors For Neuron Ranking}
\label{sec:predictor_design}

When deploying a large language model, for a given input, we will not have access to the activations or the gradients. Thus, calculating the \texttt{L2Norm} or \texttt{plainact} criterion is not possible. However, it is possible to create a calibration dataset of inputs and their corresponding \texttt{L2Norm} or \texttt{plainact} for each head and neuron. Such a dataset can be used to train a predictor, which can take the input and predict the sparsity pattern of the model at deployment. Sparsity prediction can reduce the end-to-end latency of transformer inference by predicting which operations can be skipped. 

We propose a method called ShadowLLM that uses the first layer's attention output to predict the sparsity pattern for the entire model. This reduces the overhead of repeatedly calling the predictor at each layer and cuts the total FLOPs of the predictor by 20\%, as shown in Table \ref{tab:flops_comparison}. ShadowLLM uses the activation of the first layer, which is not pruned, to predict the sparsity pattern for subsequent layers.

We also explore a \textit{Full Sequence ShadowLLM}, which uses a small transformer to take in the entire input token embedding and predict the sparsity pattern, allowing pruning of the first transformer layer as well. However, the \textit{Full Sequence ShadowLLM} requires an additional $2(2E^{2} + EL^{2})$ FLOPs, making it as costly as running an entire dense attention layer and impractical due to the high computational cost. 

\begin{table}[t]
        \centering
        \resizebox{0.90\columnwidth}{!}{%
            \begin{tabular}{l|r}
            \toprule
            \textbf{Predictor} & \textbf{FLOPs Equation}  \\
            \midrule
            DejaVu & \( N  \left(E  p_1 + p_1  (H + F)\right) \) \\
            ShadowLLM & \(\left(E p_1 + p_1  (N  (H + F))\right) \) \\
            \midrule\midrule
            \textbf{Model} & \textbf{ShadowLLM FLOPs Reduction} \\
            \midrule
            OPT-1.3B & 19.11\% \\
            OPT-30B & 19.55\% \\
            OPT-175B & 19.76\% \\
            \bottomrule
            \end{tabular}
        }
        \caption{For a transformer with E embedding dimension, N layers, H heads, F FFN neurons per layer, ShadowLLM uses (N-1)Ep$_{1}$ fewer FLOPs, where p$_{1}$ is the predictor hidden dimension. The table also shows the percentage improvement in predictor FLOPs for ShadowLLM vs. DejaVu for different models.}
        \label{tab:flops_comparison}
\end{table}

DejaVu employs a two-layer MLP, taking the activation from the final token at every alternating layer and predicting the sparsity of the next layer. 
A significant portion of the complexity of the DejaVu system arises from its asynchronous look-ahead predictor which can be expensive in wall clock time despite aggressive optimizations within DejaVu. 
The predictor itself only takes 2\% of the total FLOPs for OPT-1.3B, but having a per-layer predictor adds significant overhead due to additional GPU kernel launches and memory bandwidth constraints, leading to an end-to-end latency increase of 25\% over static sparsity (same sparsity but fixed, without a predictor).

DejaVu's per-layer approach to pruning, where a fixed sparsity is enforced per layer, can be sub-optimal as true contextual sparsity should be independent of layers. To study our proposed predictor designs in a contextual-sparsification setting, we evaluate the Spearman-$\rho$ (rank correlation coefficient) between the relative importance order of neurons and heads given by the predictor, and the relative importance order given by the pruning criterion. 
Additionally, this is done on a global head-ranking task. From Figure \ref{fig:predictor_design_test}, we see that DejaVu-style layer-wise predictors are not trained for global pruning. We find that \textit{Full Seq. ShadowLLM} performs similarly to ShadowLLM, but with a significant increase in overall FLOPs.

\begin{figure}[t]
    \centering
    \includegraphics[width=\columnwidth]{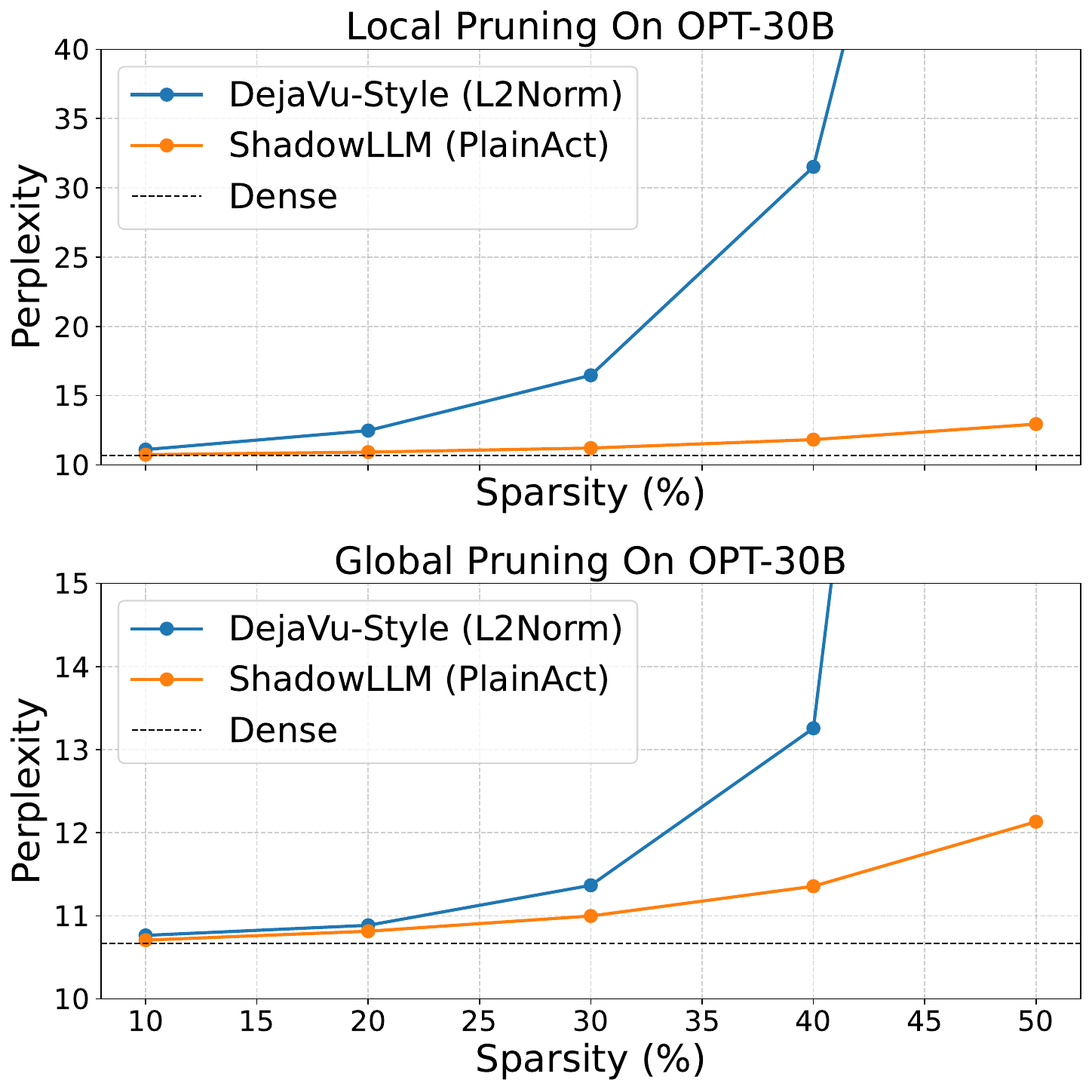}
    \caption{Comparison of the DejaVu-style predictor trained on a magnitude-based metric (\texttt{L2Norm}) with ShadowLLM on the best pruning criteria (\texttt{plainact}) on WikiText2. In both local and global settings, ShadowLLM performs well due to better pruning criteria.}
    \label{fig:opt30b_basepplx}
\end{figure}


To analyze the ability of ShadowLLM predictors to assess neuron importance in a global and local (per-layer) setting, we train ShadowLLM predictors for \texttt{plainact} across all seven down-stream tasks in the 5-shot setting, with a per-layer (local) output normalization scheme. We then evaluate the WikiText2 perplexity \footnote{For effective context-sparsity evaluation, perplexity calculations are performed on a per-document basis, differing from standard concatenation methods; see Section \ref{sec:pplx_diff} for details.} of the OPT-1.3B model as we increase sparsity for both the global and local (per-layer) pruning strategies. For local pruning, every layer achieves the target sparsity, and relative importance are only compared intra-layer. 
In Figure \ref{fig:loc_glob_1.3b}, we find that ShadowLLM is able to preserve perplexity in both global and local cases. However, we find that global pruning generally performs better than per-layer pruning. This can be attributed to the fact that some layer heads are more important than others, and forcing equal pruning ratios for all layers may cause over-parameterization in some layers and more important head to be pruned out from an under-parameterized layer.

\section{Evaluation}

We find that the activation-gradient based pruning criteria that we use in ShadowLLM are effective for downstream evaluation tasks as well as perplexity. Further, we demonstrated in Section \ref{sec:predictor_design} that ShadowLLM can predict the sparsity pattern for the entire LLM given just the input to the first layer. 
We find that the accuracy and \textit{predictability} trade-off is excellent for the \texttt{plainact} criterion, whereas other criteria were harder to learn due to outliers and high variance. In this section, we evaluate the effectiveness of combining ShadowLLM predictor design with the \texttt{plainact} criterion, compared to our implementation of DejaVu-style\footnote{To enable comparisons across pruning criteria, we have implemented our own DejaVu-style predictor in  \href{https://github.com/abdelfattah-lab/shadow_llm/}{ShadowLLM}.} predictors, trained on a magnitude based pruning criteria.  

\begin{figure}[t!]
        \centering
        \includegraphics[width=0.95\columnwidth]{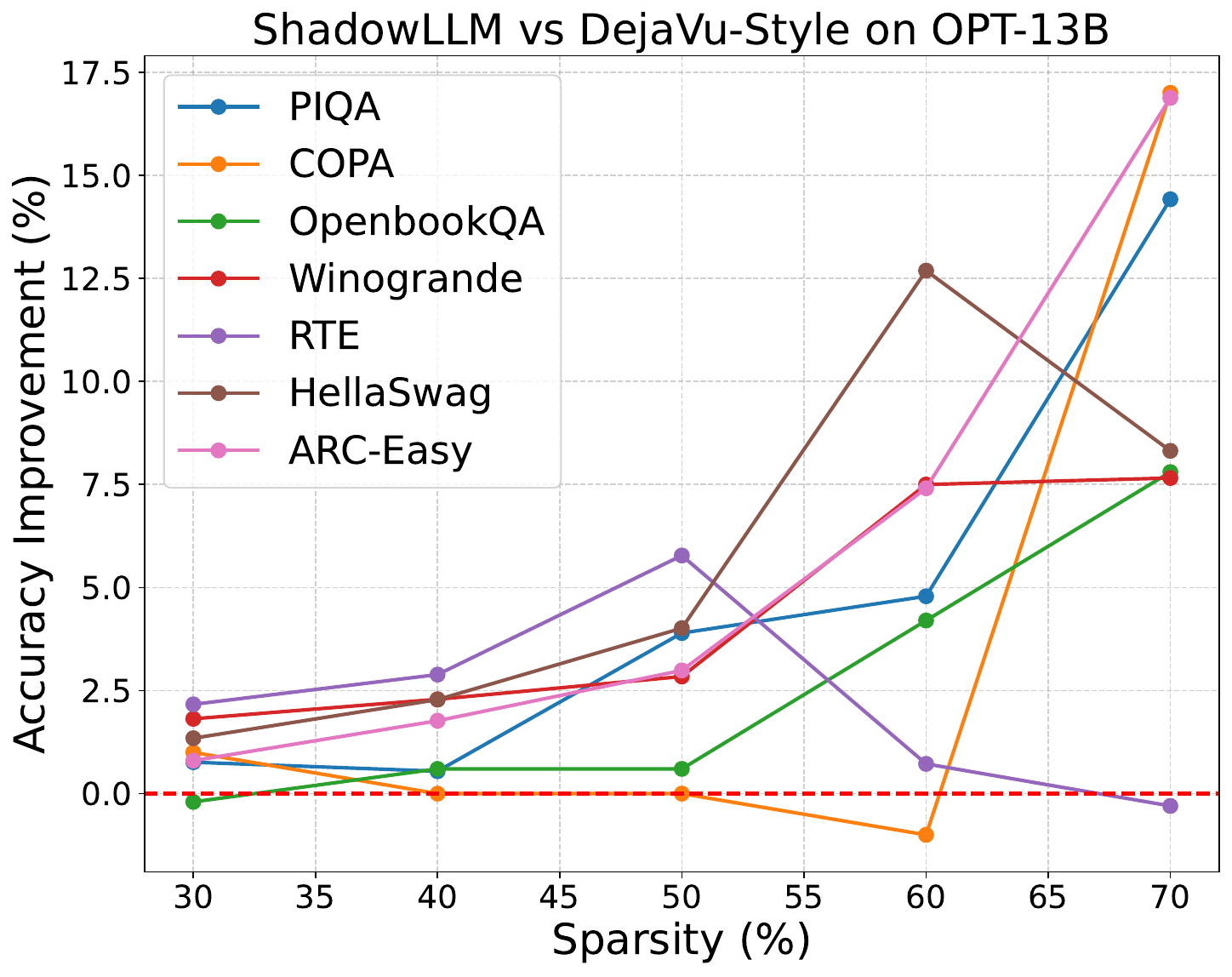}
        \caption{Consistent accuracy improvement of ShadowLLM over DejaVu across seven downstream eval tasks in the zero-shot shot setting.}
        \label{fig:downstream_13b_eval}
\end{figure}


\begin{table}[t]
    \centering
    \resizebox{0.90\columnwidth}{!}{%
        \begin{tabular}{l|r|r}
        \toprule
        \textbf{Method} & \textbf{Latency (ms)} & \textbf{Accuracy (\%)} \\
        \midrule
        Static & 2014 & 55.34 \\
        Dense & 2609 & 58.32 \\
        DejaVu & 2981 & 59.28 \\
        \textbf{ShadowLLM} & \textbf{2562} & \textbf{61.19} \\
        \bottomrule
        \end{tabular}
    }
    \caption{Latency and accuracy comparison of different methods at 50\% sparsity. Average zero-shot accuracy across 7 downstream tasks reported on OPT-13B.}
    \label{tab:latency_accuracy_comparison}
\end{table}

\begin{figure*}[ht!]
    \centering
    \begin{minipage}{0.49\textwidth}
        \centering
        \includegraphics[width=\columnwidth]{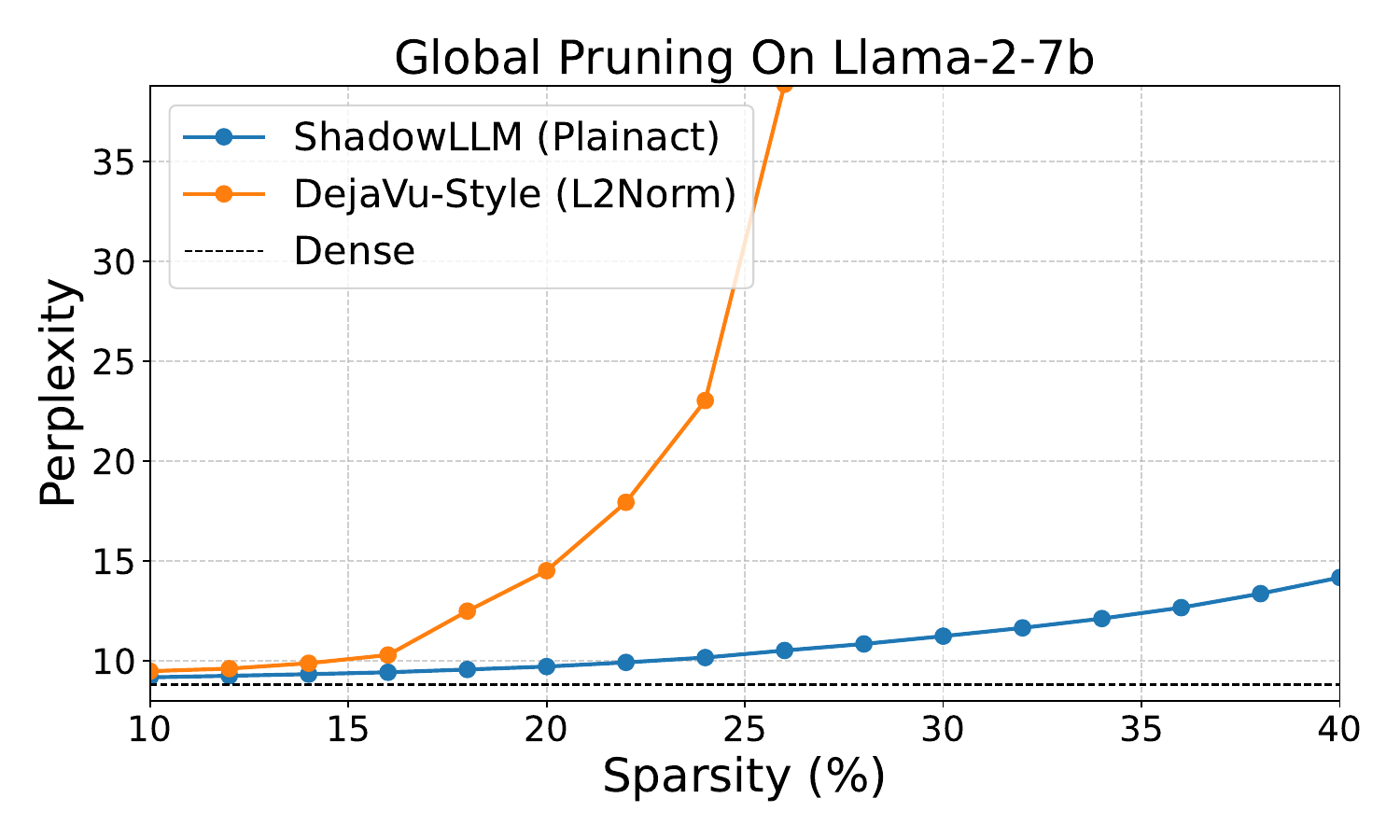}
        \caption{Gradient-informed criteria (\texttt{Plainact}) improves global pruning on Llama-2-7b, resulting in an end-to-end perplexity improvement on WikiText2.}
        \label{fig:llama_glob_pruning}
    \end{minipage}
    \hfill
    \begin{minipage}{0.49\textwidth}
        \centering
        \includegraphics[width=\columnwidth]{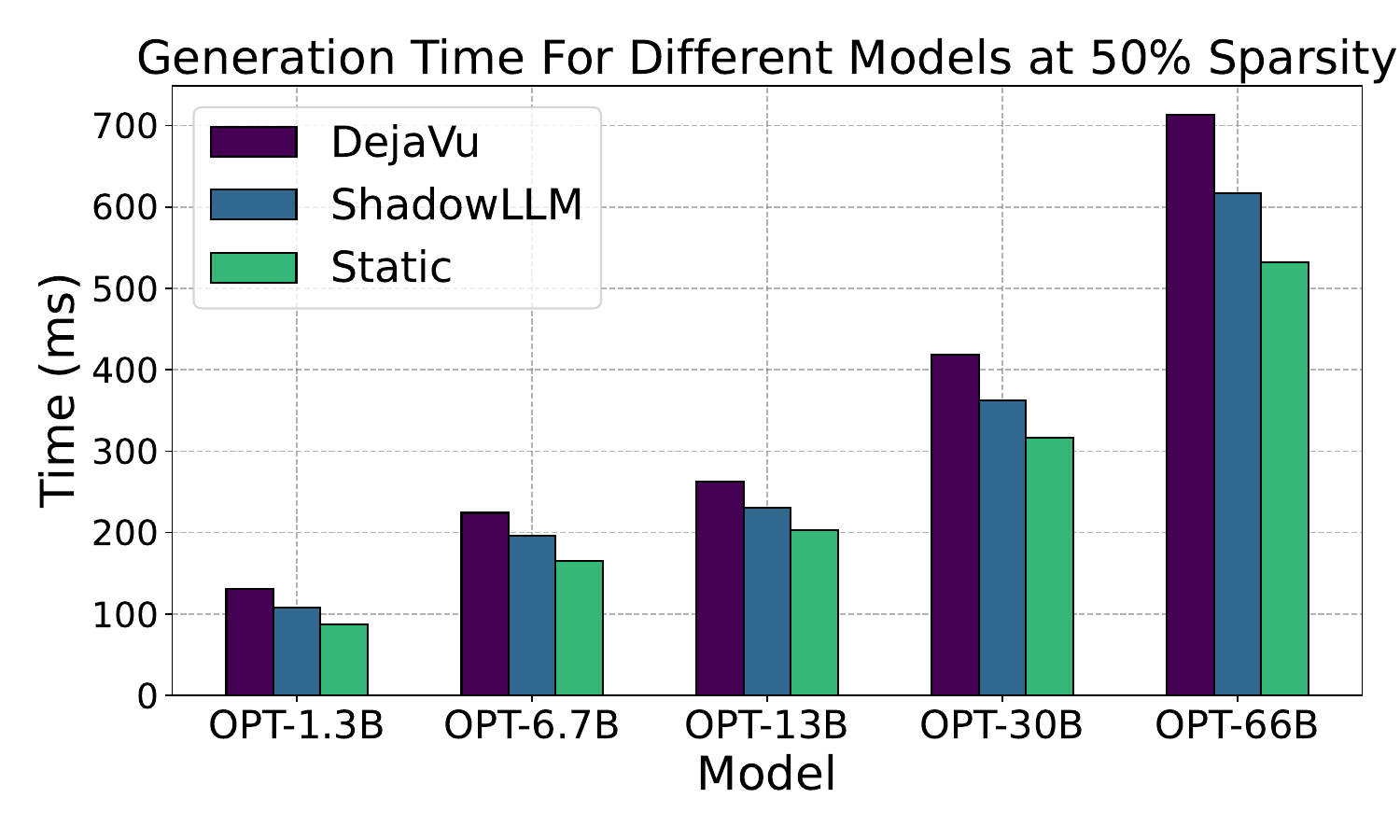}
        \caption{Average time per-inference with prompt length = 128 and generation length = 128 across model sizes. Sparsity is around 50\%.}
        \label{fig:pred_perf_scaling}
    \end{minipage}
\end{figure*}
\begin{figure*}[ht!]
    \centering
    \begin{minipage}{0.49\textwidth}
        \centering
        \includegraphics[width=\columnwidth]{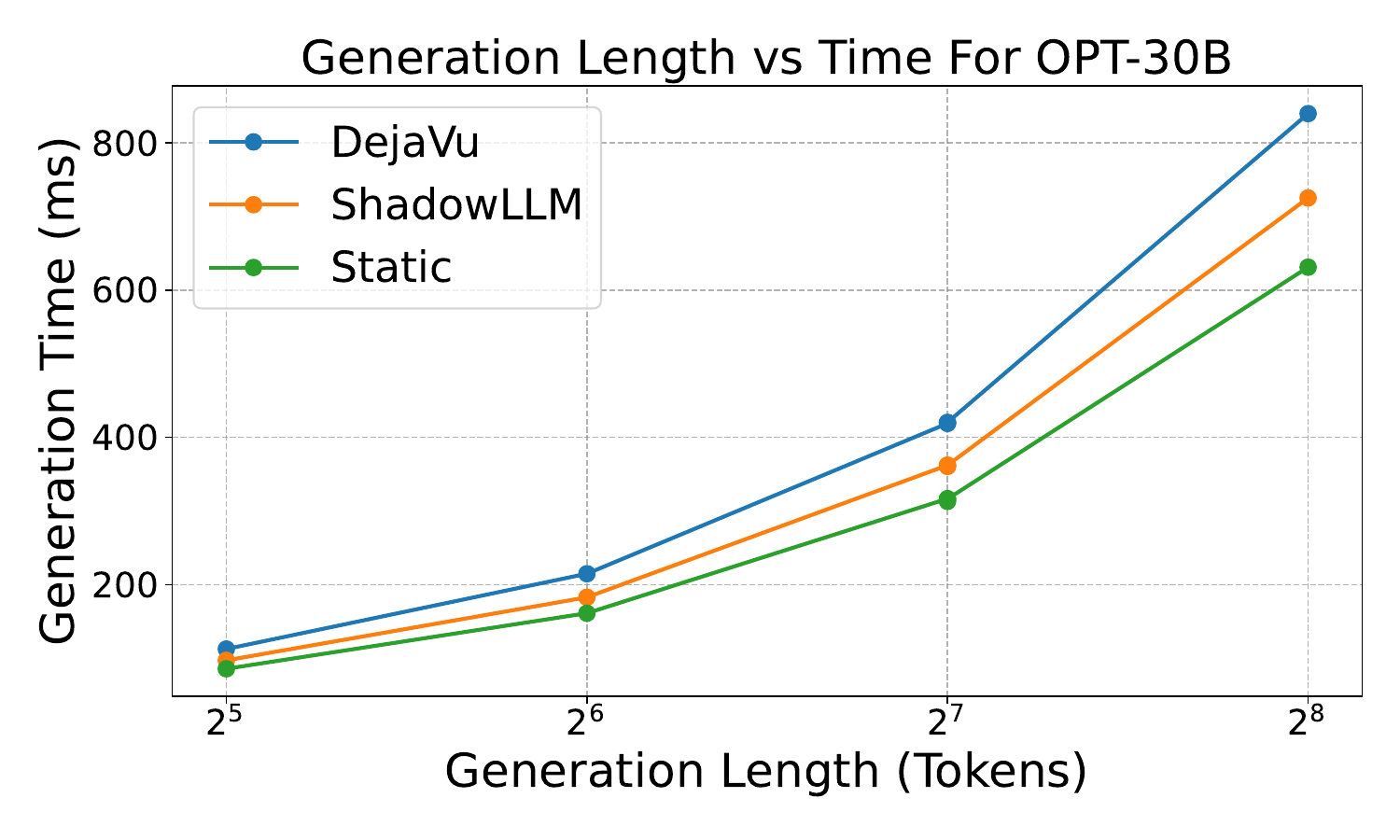}
        \caption{Average generation time on OPT-30B with prompt length = 128 as generation length increases. Sparsity is around 50\%.}
        \label{fig:pred_genlen_scaling}
    \end{minipage}
    \hfill
    \begin{minipage}{0.49\textwidth}
        \centering
        \includegraphics[width=\columnwidth]{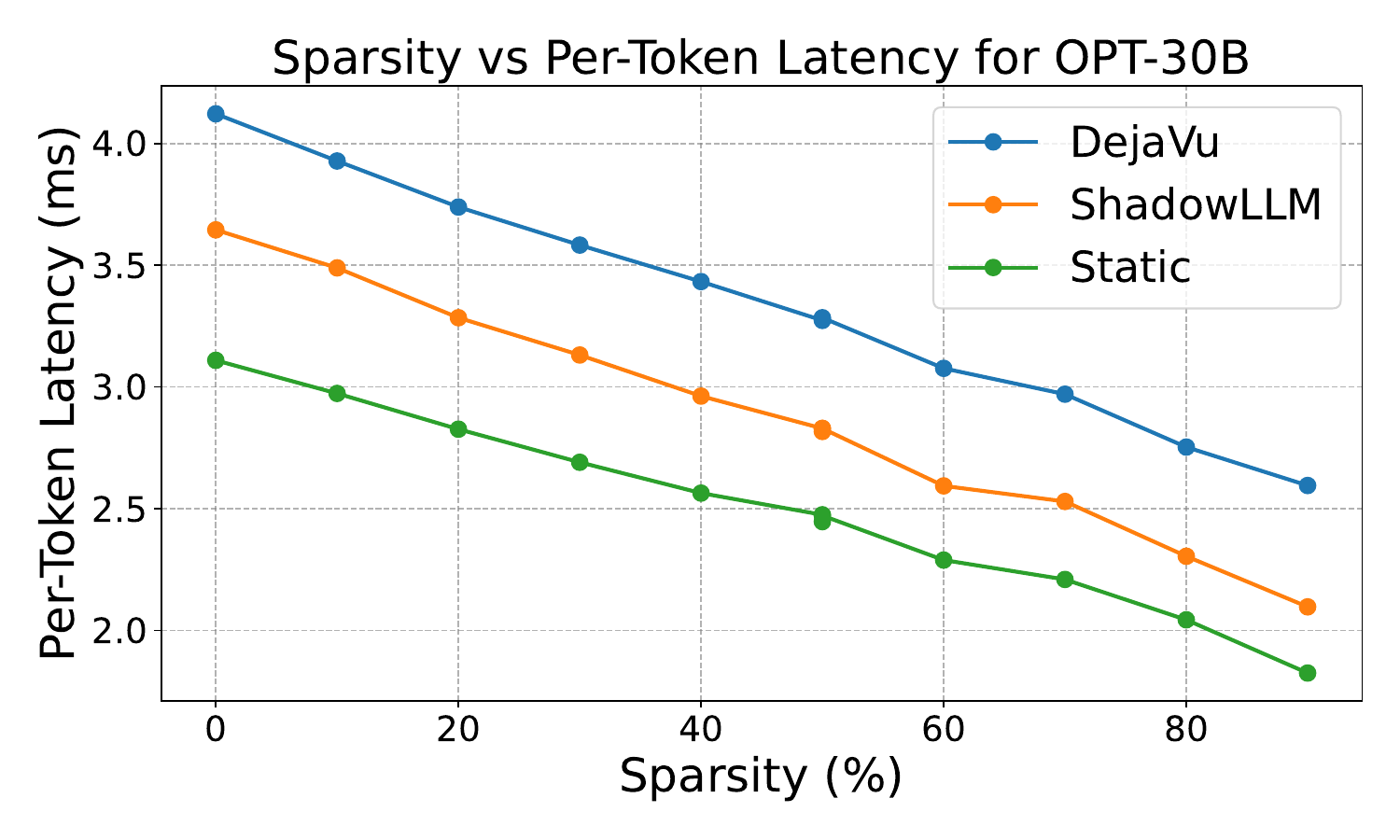}
        \caption{Per-token latency of OPT-30b with prompt length = 128 and generation length = 128 as sparsity increases.}
        \label{fig:perf_sparsity}
    \end{minipage}
\end{figure*}
\subsection{Experimental Setup}
We evaluate the perplexity for the WikiText2 ~\cite{merity2016pointer} language modeling dataset, and accuracy on 7 few-shot downstream tasks: PIQA~\cite{bisk2020piqa}, COPA~\cite{gordon2012semeval}, OpenBookQA~\cite{mihaylov2018can}, Winogrande~\cite{sakaguchi2019winogrande}, RTE~\cite{giampiccolo2007third}, HellaSwag~\cite{zellers2019hellaswag}, and ARC-Easy~\cite{clark2018think} with lm-eval-harness~\cite{eval-harness}. 

Our ablation studies to identify good pruning criteria, as well as test the efficacy of predictors is conducted on OPT-1.3B. Further, local and global pruning strategies are tested on OPT-13B and OPT-30B, and global pruning on Llama-2-7b \cite{touvron2023llama2openfoundation}. Our downstream evaluation across seven tasks is reported on OPT-13B.

\subsection{Model Quality}

In Figure \ref{fig:opt30b_basepplx}, we train the DejaVu-style and ShadowLLM predictors on their respective pruning criterion (\texttt{L2Norm} and \texttt{plainact} respectively) on 2720 input-output examples across 7 downstream tasks in the five-shot setting. The perplexity is evaluated in a local and global pruning setting on WikiText2. 

Global pruning enables better model quality - sparsity trade-off. Figure \ref{fig:llama_glob_pruning} compares the perplexity-sparsity trade-off on Llama-2-7b model, with \texttt{Plainact} significantly improving perplexity. Further, in Figure \ref{fig:downstream_13b_eval} we evaluate OPT-13B by training the ShadowLLM and DejaVu-style predictors in the same setting, and doing downstream evaluation in the zero-shot setting across seven tasks. We show that there is a consistent accuracy improvement across tasks, attributed to better pruning criteria.

We also validate these findings on OPT-13B in Figure \ref{fig:global_13b_pruning} in the global pruning setting. These improvements are largely due to an improved pruning criterion, emphasizing the importance of pruning criteria that go beyond magnitude based strategies. From Table \ref{tab:latency_accuracy_comparison} we see that ShadowLLM with the \texttt{plainact} metric delivers 14\% lower latency with 1.91\% higher accuracy than DejaVu-style predictor. 




\begin{figure*}[ht!]
    \centering
    \begin{minipage}{0.49\textwidth}
    \includegraphics[width=\columnwidth]{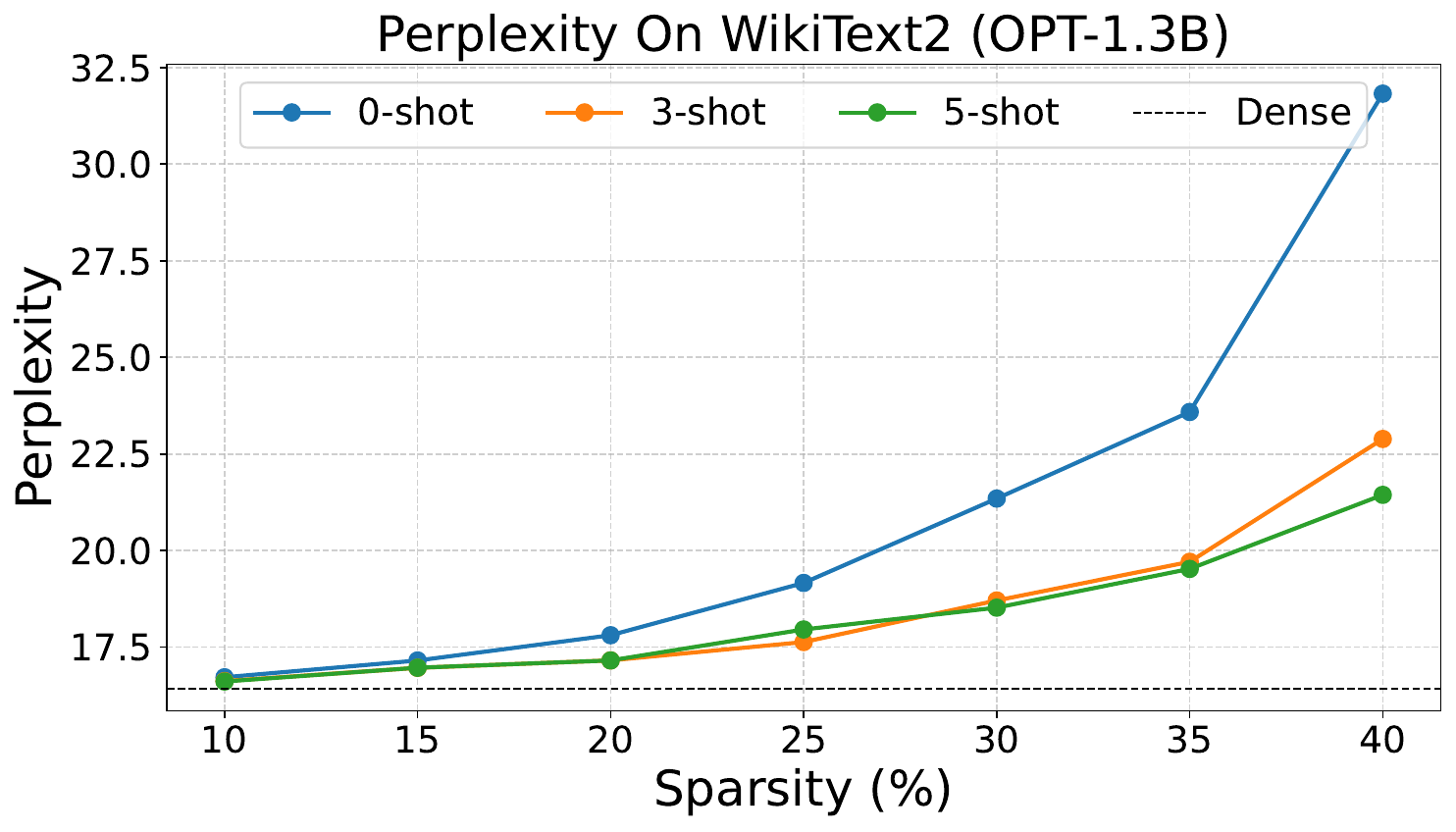}
    \caption{Calculating pruning criteria in a few-shot setting improves its ability to identify important heads and neurons.}
    \label{fig:geomean_perplexity_fewshot}
    \end{minipage}
    \hfill
    \begin{minipage}{0.49\textwidth}
        \includegraphics[width=\columnwidth]{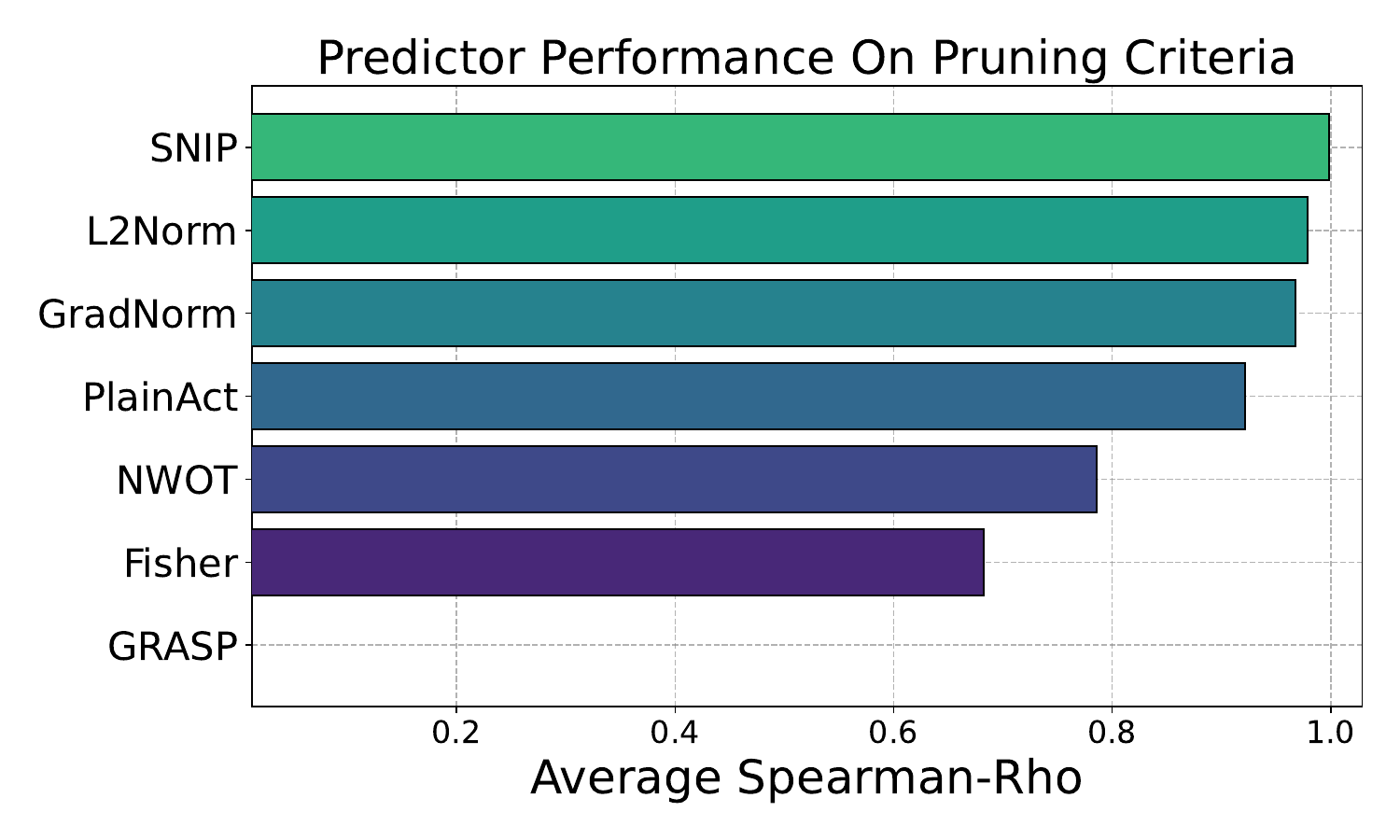}
        \caption{\texttt{plainact} is a good pruning criterion, and is also easy to learn. \texttt{grasp} has $\approx 4\times$ more outliers in proxy scores, making prediction more difficult.}
        \label{fig:predictor_target_test}
    \end{minipage}
\end{figure*}

\subsection{Performance}
DejaVu-style predictors can also be trained on better pruning criteria (\texttt{plainact}), giving improvements in accuracy. However, a single predictor can model these criteria and also offer improved end-to-end latency due to early prediction. It is also easier to integrate, without concerns for continuous pipelining and scheduling of a layer-wise predictor. In this section, we investigate the performance improvement ShadowLLM delivers by simplifying the DejaVu sparsity predictor implementation, and compare with DejaVu~\cite{liu2023deja} 

DejaVu implements hardware-efficient sparsity acceleration, which employs kernel fusion and memory coalescing to reduce overhead of the sparse matrix-vector multiply. These techniques already yield a 2$\times$ speed-up over prior SoTA FasterTransformer implementations. However, the inter-leaved sparsity predictors have a significant overhead, leading to a performance degradation of over 25\% with only a 2\% increase in total FLOPs. 

We implement the ShadowLLM predictor along with the prior enhancements introduced by DejaVu and conduct our performance experiments on up to 4 A100 GPUs. In Figure \ref{fig:pred_perf_scaling}, we measure the end-to-end time taken for a generation length of 128 tokens at 50\% sparsity and observe an average 16.2\% improvement over DejaVu. Figure \ref{fig:pred_genlen_scaling} shows a consistent improvement in generation time as output tokens increase. Further, Figure \ref{fig:perf_sparsity} shows that ShadowLLM is on average 21.25\% faster than DejaVu in the decode phase specifically. Finally, we profile model sizes from 1.3B to 66B, observing up to a 21.3\% improvement in time per-inference. 

\begin{figure*}[ht]
    \centering
    \begin{minipage}{0.49\textwidth}
        \centering
        \includegraphics[width=\textwidth]{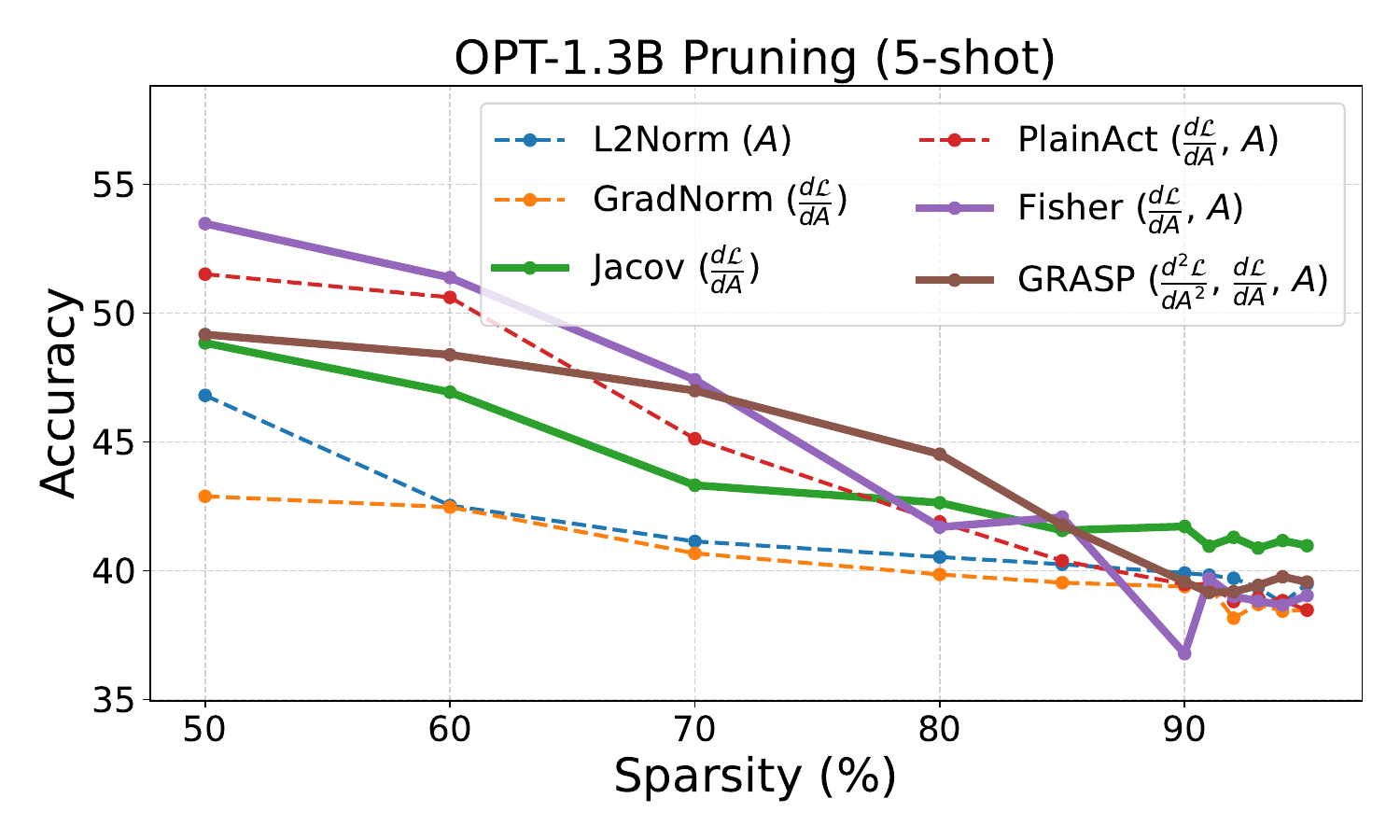}
        \caption{For every criterion, the corresponding aggregated neuron importance is used to conduct static pruning of the LLM at test time, and the average accuracy is reported.}
        \label{fig:geomean_aggr_zcp}
    \end{minipage}
    \hfill
    \begin{minipage}{0.49\textwidth}
        \centering
        \includegraphics[width=\textwidth]{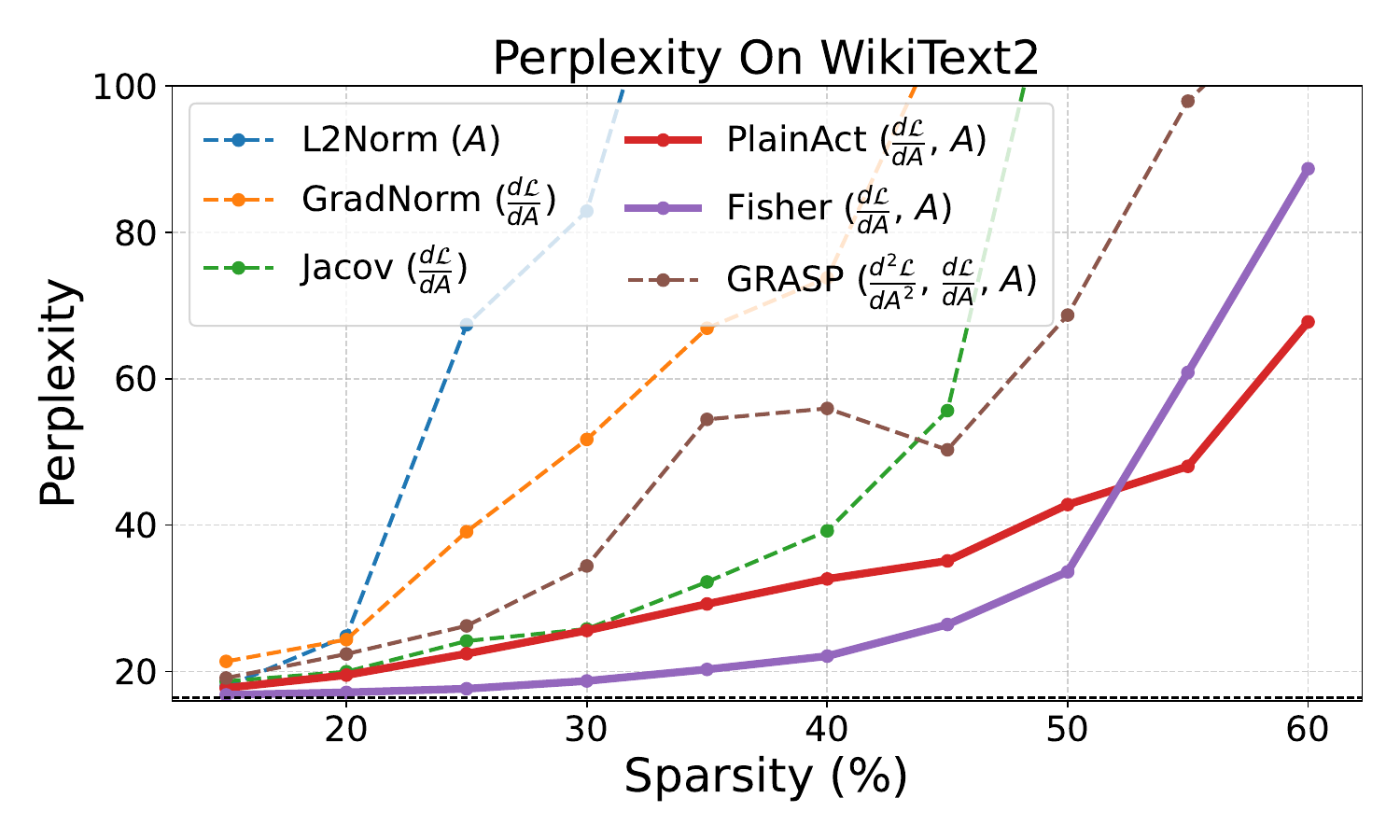}
        \caption{The aggregate importance score per neuron is used to conduct static pruning of the LLM. \texttt{fisher} and \texttt{plainact} preserve model quality better than other criteria. Dashed black line is dense baseline.}
        \label{fig:wikitext_pplx_optsmall}
    \end{minipage}
\end{figure*}




\section{Analysis}
\label{sec:analysis}

\subsection*{Overview of Pruning Criteria}

In Section \ref{sec:pruning_criterion}, we categorize pruning criteria into five primary methods: Activation Methods, First-Order Gradient (Jacobian) Methods, Activation + Jacobian Methods, OBD-style Hessian Methods, and Sensitivity-Based Methods. 

We begin by looking at activation magnitude based pruning methods akin to \cite{frankle2018lottery, han2015learning}. One such criterion, the \texttt{L2Norm} of the $k^{\text{th}}$ attention head and FFN neuron is simply $|| A_{l,k} ||_2$. More advanced methods that use gradients may provide better information about neuron importance. \texttt{GradNorm} of the $k^{\text{th}}$ attention head and FFN neuron is defined simply as $|| \frac{\partial \mathcal{L}}{\partial A_{l,k}} ||_{2}$ and $|| \frac{\partial \mathcal{L}}{\partial \theta_{l,k}} ||_{2}$ respectively. In our analysis, we found that methods that combine both the activation and Jacobian (Gradient) information perform the best. The \texttt{plainact} criterion adapted from~\cite{Bansal2022RethinkingTR, molchanov2016pruning} can be defined as $||A_{l,k} \cdot \frac{\partial \mathcal{L}}{\partial A_{l,k}}||_{1}$ and $||\theta_{l,k} \cdot \frac{\partial \mathcal{L}}{\partial \theta_{l,k}}||_{1}$ respectively. Similar to \texttt{plainact} the \texttt{fisher} criterion can be defined as $\langle (A_{l,k} \cdot \frac{\partial \mathcal{L}}{\partial A_{l,k}})^{2} \rangle$ and $\langle (\theta_{l,k} \cdot \frac{\partial \mathcal{L}}{\partial \theta_{l,k}})^{2} \rangle$ respectively, denoting a similar criterion but aggregated in a different manner. 

The \texttt{grasp} criterion approximates the change in gradient norm, which requires the Hessian $H$ and is calculated as $||-(H_{l,k} \cdot \frac{\partial \mathcal{L}}{\partial A_{l,k}}) \odot A_{l,k}||_{1}$. This OBD-style Hessian method ~\cite{wang2020picking} worked well in downstream-evaluation tasks, but did not deliver good perplexity. 

NASWOT~\cite{mellor2021neural} introduces a sensitivity based method called \texttt{jacov}. The \texttt{jacov} criterion measures the covariance of the Jacobian matrices across a mini-batch of data. \texttt{epenas}~\cite{Lopes_2021} follows the same principles as \texttt{jacov}. Naturally, as \texttt{jacov} rely on aggregated Jacobian matrices over a batch of data, this criterion cannot trivially exist for input-dependent (contextual sparsity) use-case. To test these criteria, we register the activations for the heads and neurons across the entire downstream task dataset, and generate a single \textit{aggregate head importance.} 

We evaluate the effectiveness of several pruning criteria by using them as metrics for removing less important heads/neurons. Our analysis includes evaluating the perplexity for the WikiText2~\cite{merity2016pointer} language modeling dataset and accuracy on 7 few-shot downstream tasks. 

\subsection*{Enhancing Pruning with Few-Shot Examples}
In Figure \ref{fig:geomean_perplexity_fewshot}, we calculate the \texttt{fisher} criteria for every neuron and head on 2720 input-output examples from the downstream tasks for the 0-shot, 3-shot, and 5-shot settings. We average the criteria for each neuron and head across these examples and evaluate WikiText2 perplexity as model sparsity is increased. The results indicate that providing more in-context examples when registering the criteria improves model quality during pruning.

\subsection*{Advantages of Gradient-Informed Criteria}
In Figure \ref{fig:geomean_aggr_zcp}, we use the task pruning criterion averaged over their respective examples for each head and neuron to do a static sparsification of the OPT-1.3B model and test it in the 0-shot setting. We report the mean accuracy across the downstream tasks for each pruning criteria. We find that \texttt{jacov} is a stable criteria to preserve model performance in the static case. However, \texttt{jacov} does not have a context-dependent equivalent, as it relies on the covariance of Jacobian matrices across examples. We evaluate these proxies in Figure \ref{fig:wikitext_pplx_optsmall}, and find that \texttt{fisher} and \texttt{plainact} preserve model quality well, with \texttt{jacov} performing worse. \texttt{jacov} might have higher task-dependence for static pruning, and does not translate to better general model quality.

\subsection*{Learning Pruning Criteria with Predictors}
While we can \textit{shadow} activation magnitudes with a predictor, we need to balance finding the best pruning criteria for assessing neuron importance, ensuring the criteria is easy to learn. To identify such a criteria, we measure each criteria for each head and neuron on 2720 input-output examples across the 7 downstream tasks in a 5-shot setting. We train our predictor to use the output of the first attention layer's last sequence index to predict per-head and neuron importance. Figure \ref{fig:predictor_target_test} reports the average Spearman-$\rho$ rank correlation on 680 input-output examples. From Figure \ref{fig:wikitext_pplx_optsmall}, we see that \texttt{fisher} delivers the best perplexity for up to 50\% sparsity, but delivers a Spearman-$\rho$ of under 0.7. Similarly, \texttt{grasp} is difficult to predict due to its high range and outliers. In contrast, we find that the \texttt{plainact} criterion is easy to predict and performs well in a contextual setting.

\section{Conclusion}

In this paper, we present ShadowLLM, a novel approach that realizes contextual sparsity in large language models by using a gradient-informed pruning criterion. We demonstrate that these criteria can be effectively modeled by a single predictor at the first layer of the LLM, eliminating the need for per-layer prediction. Our findings, validated on models with up to 30 billion parameters, show that relatively small predictors can model contextual sparsity in LLMs. This approach, combining an improved pruning criterion with an early predictor, enables over 15\% improvement in accuracy without a latency trade-off and a 20\% improvement in performance across different model sizes.

\section*{Limitations}
In this paper, we work towards significantly simplifying the predictor design, and study several pruning criteria. However, our study is limited to smaller models, up to 30B parameters on only OPT style models. Further, criteria like \texttt{nwot} are designed for the ReLU activation function, which may not be directly applicable to attention maps. We profile these for completeness regardless, however, more research in pruning criteria is needed. Finally, we train predictors on less than 10000 input-output examples, more examples may enable better sparsity pattern modeling. 

\section*{Acknowledgements}
This material is based upon work supported by the National Science Foundation under Grant No. 2339084, in addition to funding by Intel Corporation. 
We would like to thank Nilesh Jain, Juan Pablo Munoz, Sameh Gobriel, and Vui Seng Chua for helpful discussions and feedback.

\bibliography{acl_latex}
\newpage
\appendix 

\section{Appendix}
\label{sec:appendix}

\subsection{Predictor Design}

\begin{table}[h]
\centering
\begin{tabular}{l l}
\toprule
\textbf{Hyper-parameter} & \textbf{Value} \\
\midrule
Hidden Layers & 1 \\
Hidden Layer Neurons & 2048 \\
Activation Function & ReLU \\
Input Dimension & Model Embedding \\
Output Dimension & Number Of Neurons \\
Number of Epochs & 100 \\
Batch Size & 32 \\
Optimizer & AdamW \\
Learning Rate & 0.001 \\
Scheduler & CosineAnnealingLR \\
Criterion & MSELoss \\
\bottomrule
\end{tabular}
\caption{Hyperparameters for DejaVu-style and ShadowLLM predictor training.}
\label{tab:hyperparameters}
\end{table}

\begin{figure}[b]
        \centering
        \includegraphics[width=\columnwidth]{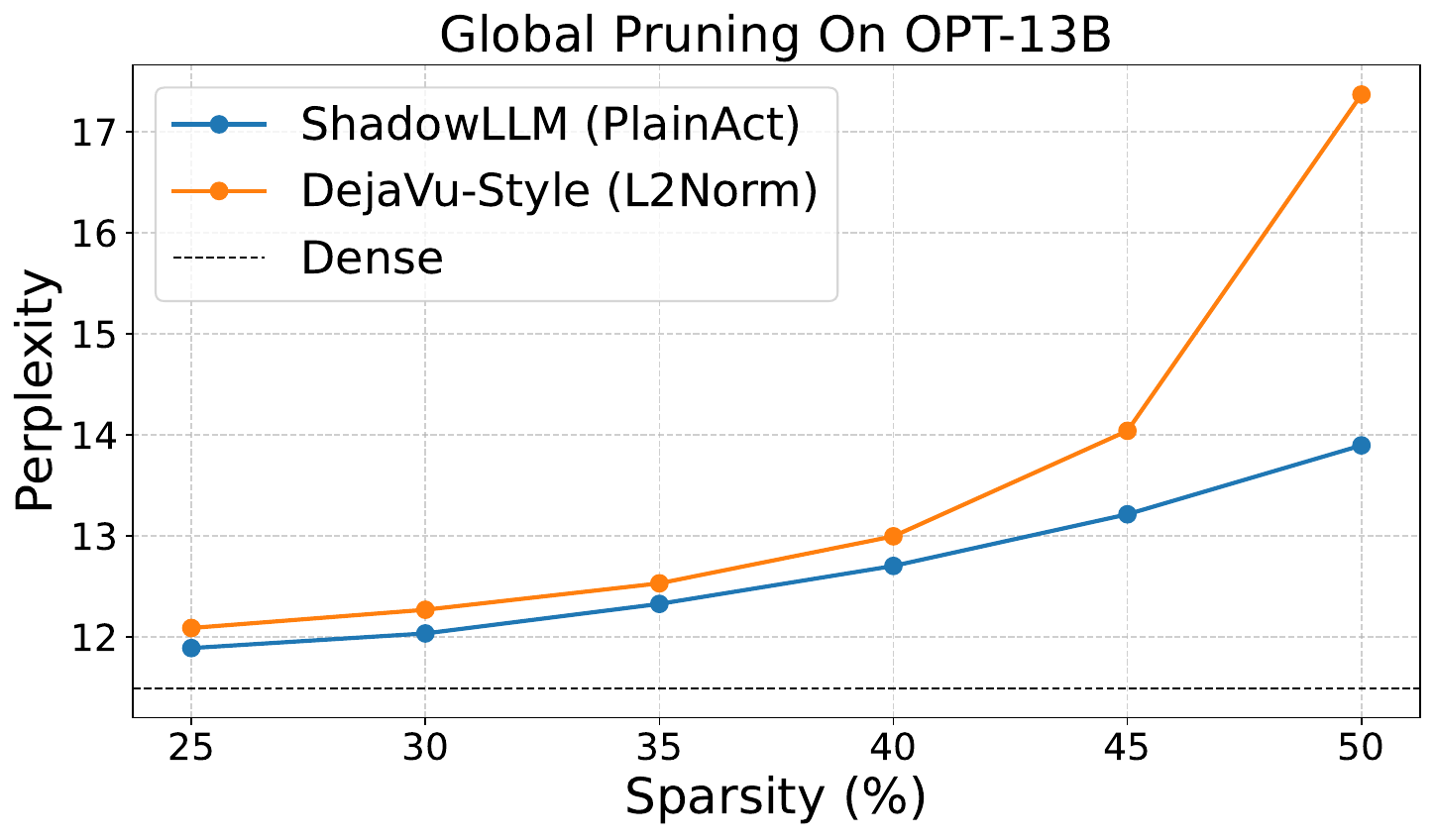}
        \caption{A better criteria (\texttt{plainact}) with the ShadowLLM predictor improves perplexity-sparsity trade-off on WikiText2.}
        \label{fig:global_13b_pruning}
\end{figure}

\subsection{Additional Pruning Criteria}

\begin{figure*}[ht]
    \centering
    \begin{minipage}{0.49\textwidth}
        \centering
        \includegraphics[width=\textwidth]{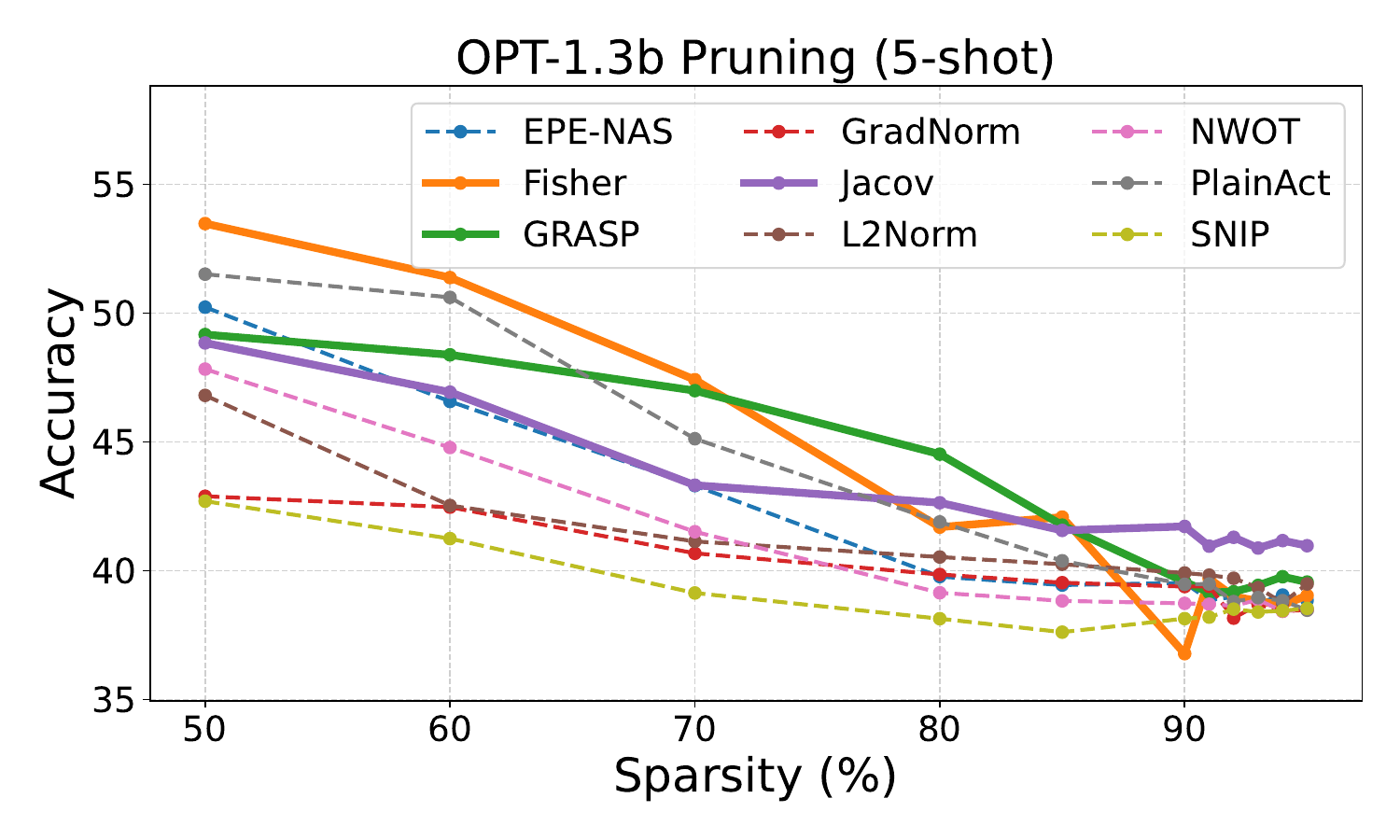}
        \caption{Each pruning criterion is measured and averaged per neuron and head over 3500 training examples in a 5-shot setting across 7 downstream tasks. For every criterion, the corresponding aggregated neuron and head importance is used to conduct static pruning of the LLM at test time. For each criterion, mean of accuracy is reported as sparsity is increased.}
        \label{fig:geomean_aggr_zcp_appx}
    \end{minipage}
    \hfill
    \begin{minipage}{0.49\textwidth}
        \centering
        \includegraphics[width=\textwidth]{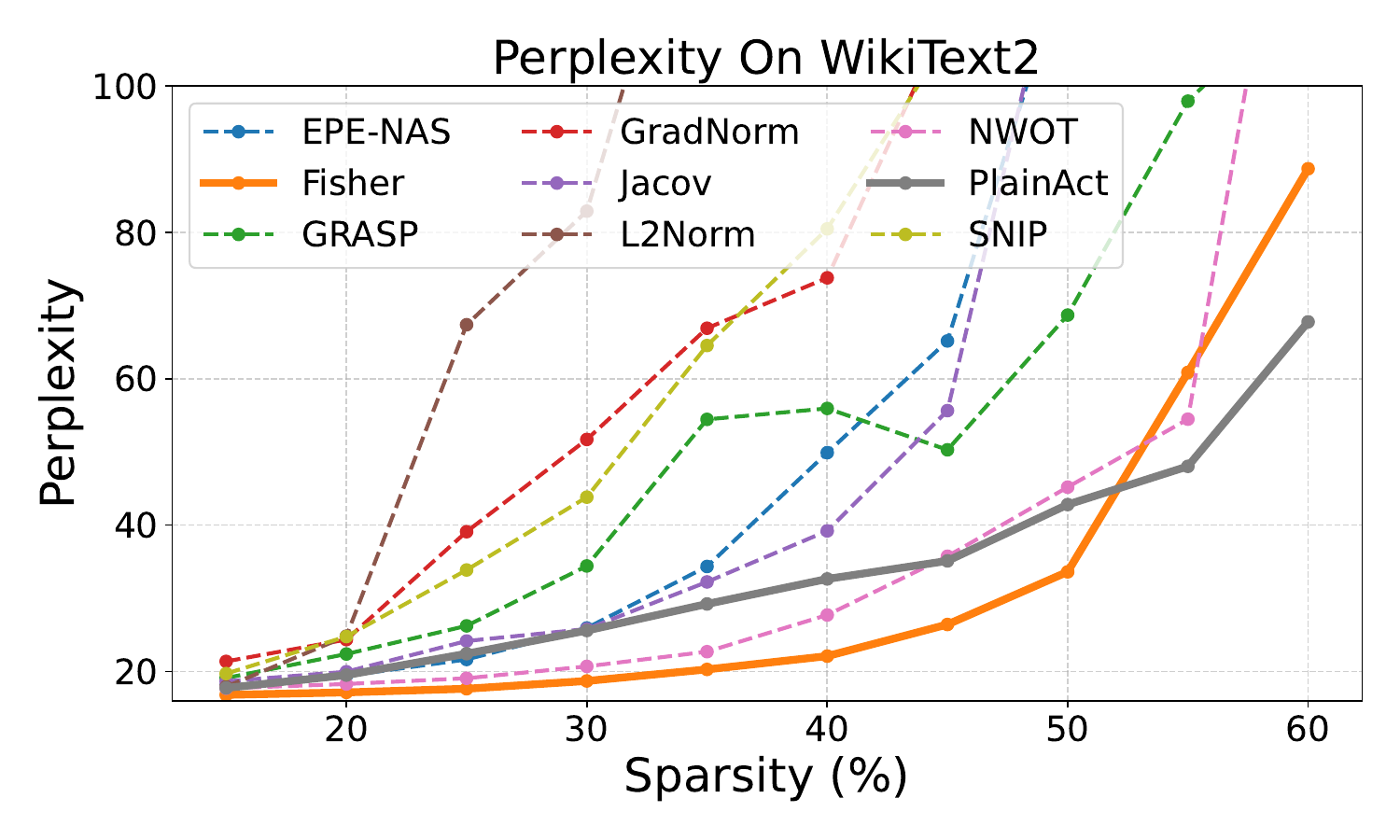}
        \caption{Each pruning criterion is measured and averaged per neuron and head over 2720 training examples in a 5-shot setting from all downstream tasks. This aggregate importance score per neuron and head is used to conduct static pruning of the LLM when testing perplexity on WikiText2. \texttt{fisher} and \texttt{plainact} preserve model quality better than other criteria.}
        \label{fig:wikitext_pplx_optsmall_appx}
    \end{minipage}
\end{figure*}
In this section, we provide a more complete view of the proxies we investigate and their results. 

While some criteria were designed for activations (\texttt{Fisher}), whereas others for weights (\texttt{snip}), we extend the pruning criteria to both activations for attention heads and weights for FFN neurons. A side-effect of this is that criteria such as \texttt{plainact} and \texttt{fisher} look similar, but are aggregated in different ways (L1 Norm versus Mean). We maintain both variants in our analysis for completeness.

Similar to \texttt{jacov}, \texttt{epenas} is also a viable method for non-contextual sparsity. \texttt{epenas} measures the intra- and inter-class correlations of the Jacobian matrices. We modify \texttt{epenas} by treating next-tokens as the class that the Jacobians are registered as. 

We also investigate sensitivity-based methods, such as \texttt{snip}~\cite{lee2018snip}, defined in Equation \ref{eq:snipdefn}, which investigates how removing a single neuron in isolation will impact the loss. 

\begin{equation}
 \texttt{snip} = \displaystyle \lim_{\varepsilon  \to 0} \left| \frac{\mathcal{L}_{\theta} - \mathcal{L}_{\theta + \varepsilon \delta_{q}}}{\varepsilon} \right|
\label{eq:snipdefn}
\end{equation}

Further, we adapt proxies from neural architecture search for neuron saliency. The NASWOT \cite{mellor2021neural} paper introduces two criteria, the first we refer to as \texttt{nwot}. \texttt{nwot} calculates the determinant of a Hamming distance-based kernel matrix, which measures the similarity of binary codes that result after the ReLU, given an input in the neural network. This uses the intuition that if two similar inputs lie within the same linear region of the network, they will be difficult to disentangle. \texttt{nwot} is defined in Equation \ref{eq:nwotdefn}.

\begin{equation}
\texttt{nwot}_{l,k} = \log\left(\frac{1}{\text{seqlen}} \sum_{i=1}^{\text{seqlen}} (1 - A^{i}_{l,k})^{2}\right)
\label{eq:nwotdefn}
\end{equation}

\begin{table}[t]
\centering
\begin{tabular}{l r r}
\toprule
\textbf{Model} & \textbf{Ours} & \textbf{Reported} \\
&  & \textbf{\cite{lin2024awqactivationawareweightquantization}} \\
\midrule
Llama2-7B & 8.82 & 5.47 \\
OPT-1.3B & 16.4 & 14.6 \\
OPT-13B & 11.5 & 10.1 \\
OPT-30B & 10.7 & 9.56 \\
\bottomrule
\end{tabular}
\caption{Comparison of our perplexity and reported perplexity for various models. Following \cite{Bansal2022RethinkingTR}, we calculate per-document perplexity, which increases model perplexity.}
\label{tab:perplexity_comparison}
\end{table}

\section{On Perplexity Calculation}
\label{sec:pplx_diff}
In our experiments, we evaluate the perplexity of language models on the WikiText-2 dataset \textbf{by computing the log-likelihood of each document individually, rather than concatenating all documents into a single continuous text stream.} Specifically, we process each document separately, calculating perplexity within the that document's context. This approach limits the context to within individual documents, without leveraging cross-document dependencies that the standard concatenation method from reference works provide (e.g. \texttt{"\textbackslash n\textbackslash n".join(wikitext\_docs['text'])}). As a result, our perplexity scores reflect the model's performance on isolated text segments, which may differ from scores obtained using the more conventional concatenated approach. While this methodology deviates from standard practice, it offers a consistent evaluation of the model's capabilities within document-level context, aligning with the setting considered in our study as well as the in-context learning literature we build our study on ~\cite{Bansal2022RethinkingTR}. We quantify this difference in perplexities in Table \ref{tab:perplexity_comparison}.

\begin{figure*}
    \centering
    \includegraphics[width=\textwidth]{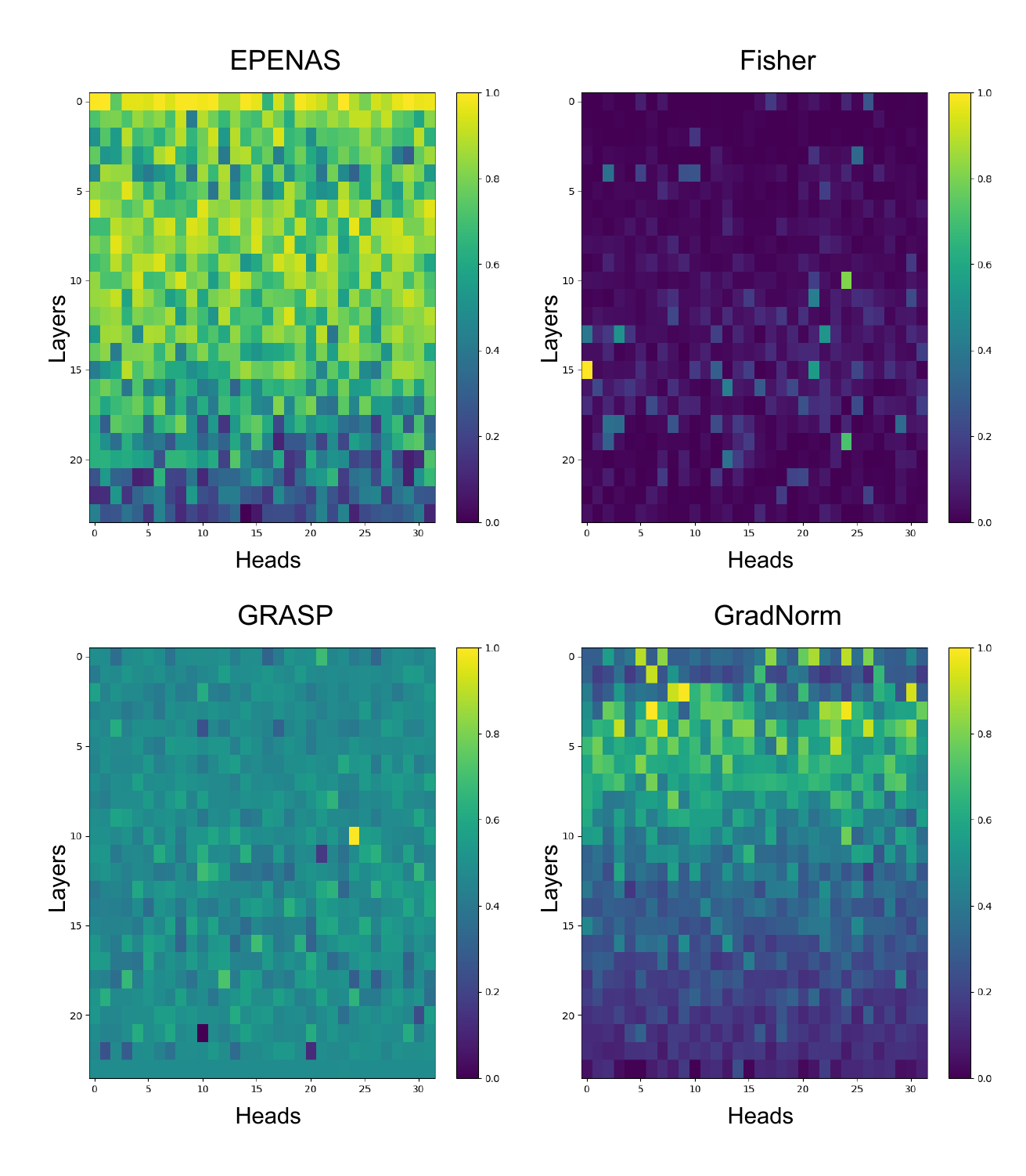}
    \caption{Aggregated pruning criteria scores per-head for the OPT-1.3B model, over the ARC-Easy training task in a five-shot setting. }
    \label{fig:aggr_corrg_score_1}
\end{figure*}

\begin{figure*}
    \centering
    \includegraphics[width=\textwidth]{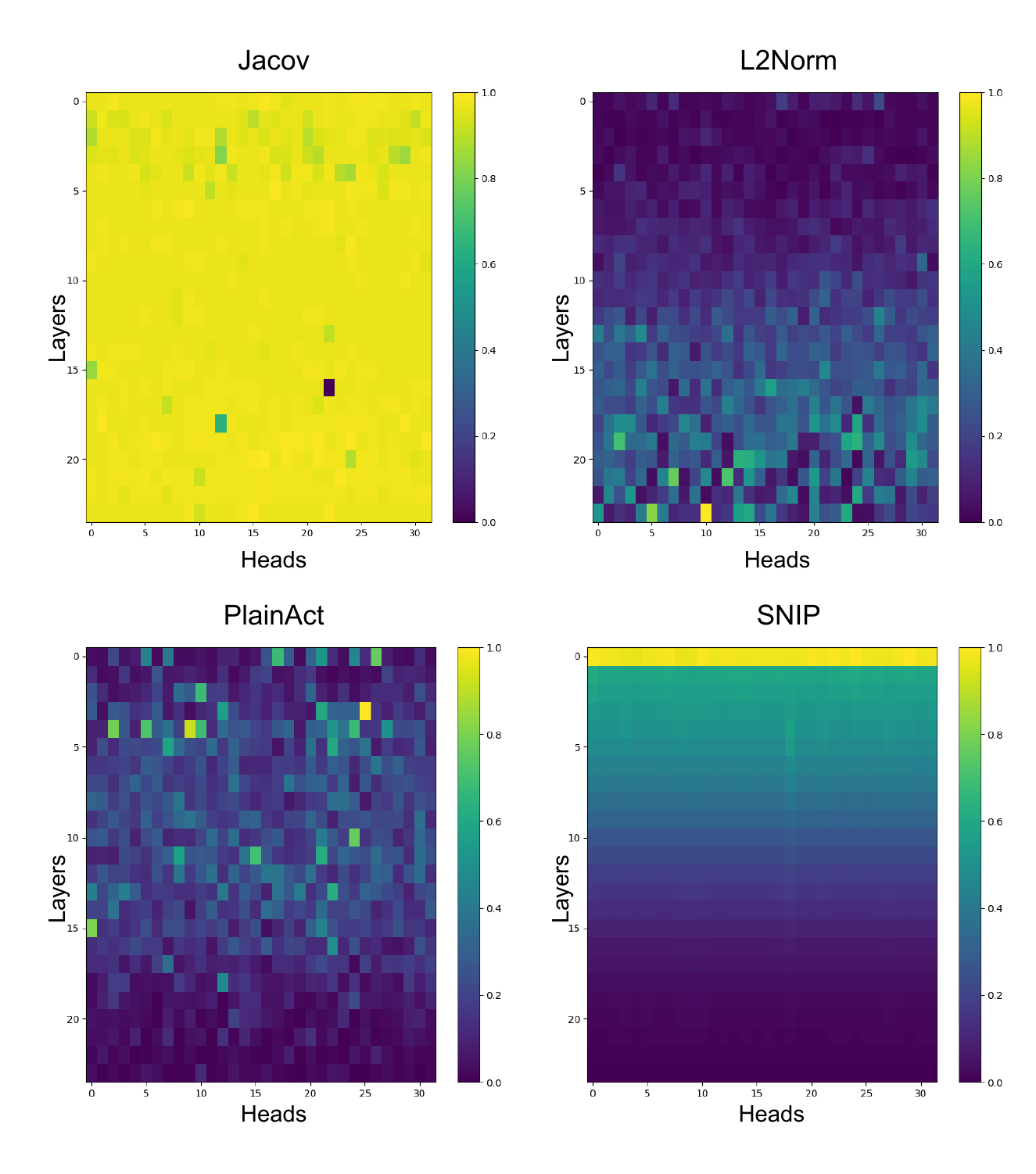}
    \caption{Aggregated pruning criteria scores per-head for the OPT-1.3B model, over the ARC-Easy training task in a five-shot setting. }
    \label{fig:aggr_corrg_score_2}
\end{figure*}

\end{document}